\pdfoutput=1

\documentclass[11pt]{article}

\usepackage{ACL2023}

\usepackage{times}
\usepackage{latexsym}

\usepackage[T1]{fontenc}

\usepackage[utf8]{inputenc}

\usepackage{microtype}

\usepackage{inconsolata}
\usepackage{xcolor}
\usepackage{multirow}
\usepackage{booktabs}
\usepackage{makecell}
\usepackage{amssymb}
\usepackage{colortbl}
\usepackage{graphicx}
\usepackage{bbding}
\usepackage{utfsym}
\usepackage{caption}
\usepackage{subcaption}
\usepackage{amsmath}
\usepackage{enumitem}

\usepackage[export]{adjustbox}

\definecolor{ao(english)}{rgb}{0.0, 0.5, 0.0}
\usepackage{pifont}
\usepackage{xcolor}
\usepackage{xspace}
\definecolor{ForestGreen}{RGB}{34,139,34}
\definecolor{BrickRed}{rgb}{.72,0,0}
\definecolor{LakeBlue}{RGB}{0,61,153}

\newcommand{\sqs}[1]{{\color{purple}{[(SQS): #1]}}}

\newcommand{\slm}{Symbol-LLM\xspace}
\newcommand{\slmi}{Symbol-LLM\textsubscript{Instruct}\xspace}
\newcommand{\slmb}{Symbol-LLM\textsubscript{Base}\xspace}

\newlength{\RoundedBoxWidth}
\newsavebox{\GrayRoundedBox}
   {\setlength{\RoundedBoxWidth}{\dimexpr#1}
    \begin{lrbox}{\GrayRoundedBox}
       \begin{minipage}{\RoundedBoxWidth}}%
   {   \end{minipage}
    \end{lrbox}
    \begin{center}
    \begin{tikzpicture}%
       \draw node[draw=black,fill=black!10,rounded corners,%
             inner sep=2ex,text width=\RoundedBoxWidth]%
             {\usebox{\GrayRoundedBox}};
    \end{tikzpicture}
    \end{center}}

\usepackage{tcolorbox}
\newtcolorbox{finding}{
colframe=black!80,
colback=icyrockblue,
fonttitle=\bfseries,
coltitle=black,
left=3pt,
right=3pt,
top=3pt,
bottom=3pt,
boxrule=0.4mm,
arc=3mm
}

\title{\protect\includegraphics[scale=.01, valign=c]{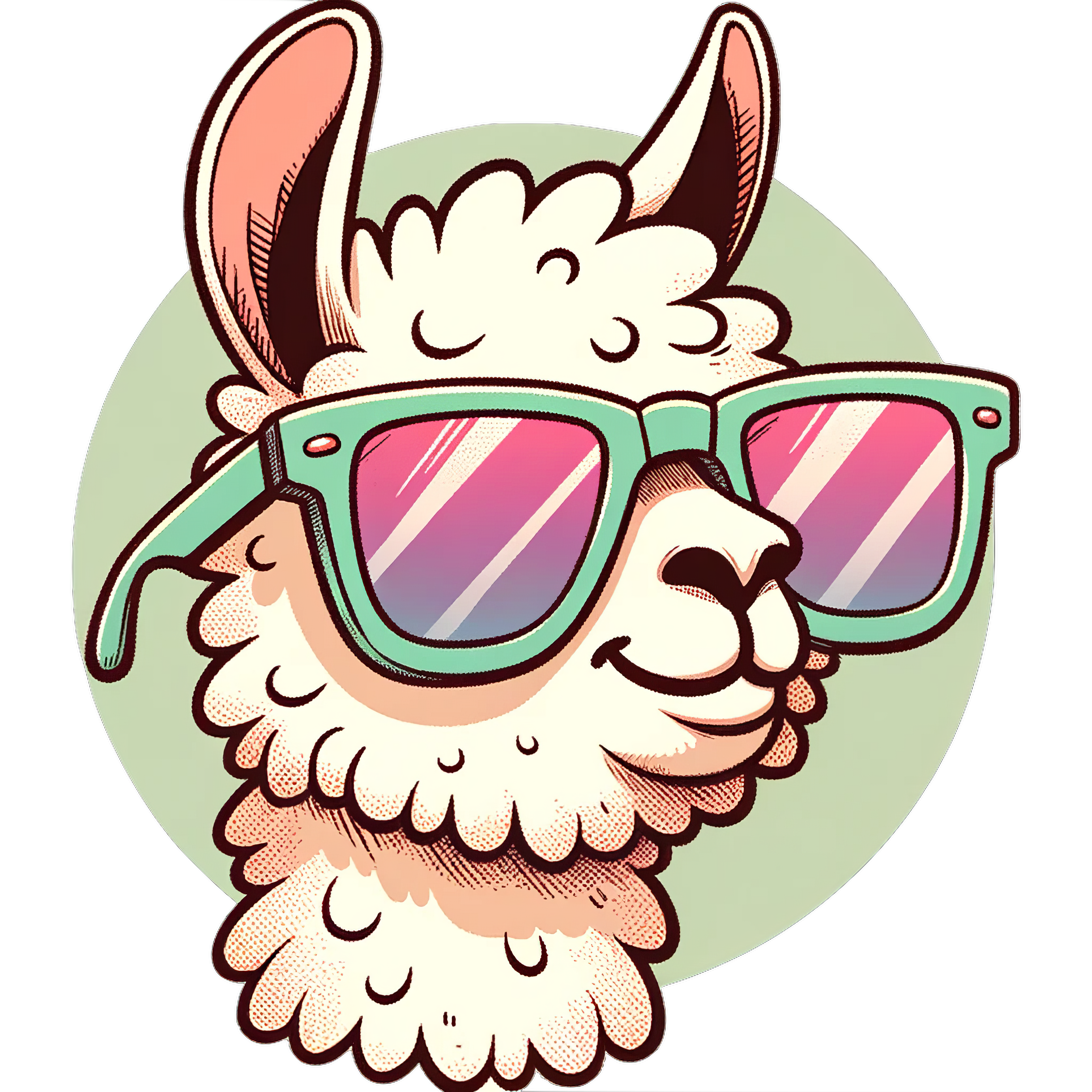} Symbol-LLM: Towards Foundational Symbol-centric Interface \\For Large Language Models}

\author{Fangzhi Xu$^{1,2}$, Zhiyong Wu$^{2}$\thanks{\, Corresponding Author.}, Qiushi Sun$^{3}$, Siyu Ren$^{4}$, Fei Yuan$^{2}$, Shuai Yuan$^{5}$, \\
            {\bf Qika Lin$^{3}$, Yu Qiao$^{2}$, Jun Liu$^{1}$}\\
        $^1$Xi'an Jiaotong University  \: $^2$Shanghai Artificial Intelligence Laboratory \\
        $^3$National University of Singapore  \: $^4$Shanghai Jiao Tong University  \\
        $^5$Hong Kong University of Science and Technology  \\
        \texttt{Leo981106@stu.xjtu.edu.cn, wuzhiyong@pjlab.org.cn, qiushisun@u.nus.edu}
        }

\begin{document}
\maketitle

\begin{abstract}
Although Large Language Models (LLMs) demonstrate remarkable ability in processing and generating human-like text, they do have limitations when it comes to comprehending and expressing world knowledge that extends beyond the boundaries of natural language(e.g., chemical molecular formula).
Injecting a collection of symbolic data directly into the training of LLMs can be problematic, as it disregards the synergies among different symbolic families and overlooks the need for a balanced mixture of natural and symbolic data. In this work, we tackle these challenges from both a data and framework perspective and introduce Symbol-LLM series models.
First, we curated a data collection consisting of 34 tasks and incorporating approximately 20 distinct symbolic families, intending to capture the interrelations and foster synergies between symbols. Then, a two-stage tuning framework succeeds in injecting symbolic knowledge without loss of the generality ability. Extensive experiments on both symbol- and NL-centric tasks demonstrate the balanced and superior performances of Symbol-LLM series models. The project page is \url{https://xufangzhi.github.io/symbol-llm-page/}.

\if
Large Language Models (LLMs) have greatly propelled the progress in natural language (NL)-centric tasks. 
Yet,
these advancements address only a subset of world knowledge.
Symbolic languages,
extending beyond the NL domain,
excel in encapsulating a broader spectrum of expressing knowledge and integrating external tools.
However, two critical challenges persist in integrating symbolic knowledge into LLMs:
First, 
the interrelations between various symbols remain unexplored. 
Second, 
achieving a balance between symbol-centric and NL-centric capabilities presents considerable difficulties.
In this work, 
we introduce Symbol-LLM series models
\footnote{We will open-source 7B and 13B model variants both with Base and Instruct versions.}, 
which handle the two problems respectively from data and framework perspectives.
From the angle of data,
we collect 34 symbolic tasks,
covering $\sim$20 different forms by 
(1) exploiting existing benchmarks;
(2) prompting LLMs; 
(3) employing our \emph{symbol-evol} strategy.
These tasks are unified to capture symbol interrelations.
In terms of framework,
a two-stage tuning framework is proposed to mitigate catastrophic forgetting. 
Specifically,
\emph{Injection} stage includes symbolic tasks to train on LLaMA-2-Chat models, 
obtaining \slmb. 
\emph{Infusion} stage combines instruction-tuning data and symbolic data to support ongoing training,
leading to the development of \slmi. 
Extensive experiments on both symbol- and NL-centric tasks demonstrate the balanced and superior performances of Symbol-LLM series models.\footnote{Work done during the internship at Shanghai AI Lab.}
\sqs{Abs v1 done, I tried to make it concise, have a check.}

\fi



\end{abstract}
\section{Introduction}


Large Language Models~(LLMs), such as GPT-series~\cite{radford2019language,brown2020language,OpenAI2023GPT4TR} and LLaMA-series~\cite{touvron2023llama,touvron2023llama2}, 
boosted the performance in various Natural Language Processing (NLP) tasks~\cite{zhao2023survey,wei2022chain,zhou2022least,yao2023tree}.
The success of these models heavily relies on natural language (NL) as the primary interface\footnote{\emph{Interface} in this paper refers to the communication between LLM and environment (i.e., external tools).} for interaction and reasoning. 
However, the NL-centric interface confines the inputs and outputs to an NL form, which can only address certain aspects of world knowledge, such as fact~\cite{bordes2015large}, commonsense~\cite{talmor-etal-2019-commonsenseqa}.







Nevertheless,
a substantial amount of abstract knowledge,
notably in areas like molecular formula (e.g., C$_6$H$_{12}$O$_6$) and first-order logic (e.g., $\mathtt{IsTriangle}(X) \rightarrow \mathtt{SumOfAngles}(X,180^{\circ})$),
is more effectively represented in symbolic forms rather than in NL.\footnote{For clarity, the world knowledge covered in this paper refers to the symbolic form of knowledge, rather than the board concept of 'knowledge' stored in LLMs.}


Compared to the NL form, the symbolic form covers a wide spectrum of scenarios and tends to be more concise and clear, enhancing its communication effectiveness~\cite{gao2023pal,qin2023toolllm}. 
In particular, 
when interacting with robots, symbolic command sequences (such as $\mathtt{PICKUP}$, $\mathtt{WALK}$) are more accurate and efficient than NL. Similarly, when using programming languages (like SQL and Python) to call external tools~\cite{gao2023pal}, expressing this structured information in NL form can be difficult. 

\if
Firstly, 
the underlying relations among various symbols remain largely unexplored. 
Through the lens of neuro-symbolic AI, 
plenty of works
~\citep[][\emph{inter alia}]{li-srikumar-2019-augmenting,DBLP:conf/aaai/BevilacquaBN21,edwards-etal-2021-text2mol}.
have been devoted to studying specific symbols.
Contemporarily, 
prevalent LLMs~\cite{roziere2023code,luo2023wizardcoder} also demonstrate abilities in tackling certain symbolic forms. 
Nevertheless,
despite the intuitive understanding that various symbolic forms share common patterns,
e.g.,
the atom unit (e.g., $\mathtt{BornIn}(\mathrm{Obama,USA)}$) in FOL is similar to function (e.g., $\mathtt{query(Paris, nwr(hotel))}$) in API calls in the form,
leveraging the interrelation among symbols is untouched.

In our work, 
we treat a wide range of symbolic tasks \textit{in a unified view},
anticipating that the interplay and mutual reinforcement among different symbols will advance the frontiers of neuro-symbolic AI.

Secondly,
it is challenging to balance the performances between NL- and symbol-centric tasks, 
considering the highly heterogeneous data will lead to catastrophic forgetting~\cite{kirkpatrick2017overcoming} when infusing symbolic knowledge into LLMs.
Recent works either ignore the drops in general capability~\cite{yang2023harnessing} or simply mix different types of data~\cite{xu2023lemur},
which heavily relies on a ``key formula'' for data split ratio. 
\fi

Despite the symbolic form offering a wealth of information, deploying LLMs directly via a symbolic-centric interface poses a significant challenge. This is largely attributed to the fact that LLMs are trained via large-scale unsupervised pre-training on extensive general text datasets, which inherently lack a symbolic foundation. The most straightforward approach to incorporating symbolic knowledge into LLMs is through fine-tuning~\cite{yang2023harnessing,xu2023lemur}. However, the format of symbolic data significantly diverges from that used during pre-training. Consequently, merely fine-tuning with large heterogeneous data can lead to catastrophic forgetting~\cite{kirkpatrick2017overcoming}.

Meanwhile, existing injection methods primarily stress on specific symbols, 
it is important to note that symbolic forms can be quite complex and vary across tasks. 
Training LLM for a specific symbolic form is both time-consuming and labor-intensive. 
Furthermore, 
treating each symbol independently often overlooks the interconnections between different symbols,
e.g., atom unit (e.g., $\mathtt{BornIn}(\mathrm{Obama,USA)}$) in first-order logic (FOL) is similar in form to function (e.g., $\mathtt{query(Paris, nwr(hotel))}$) in API calls .


Upon this observation,  we conduct a comprehensive collection of 34 text-to-symbol generation tasks with $\sim$20 standard symbolic forms introduced with instruction tuning format. The symbolic data comes from three sources: 
(1) 88.3\% of the data was collected from existing benchmarks. 
(2) 5.8\% of the data was prompted by LLMs. Compensating for the natural absence of symbolic representations in some NL-centric tasks, prompting powerful LLMs can generate more novel text-to-symbol pairs.
(3) 5.9\% of data was generated by introducing the \emph{Symbol-evol} strategy, with replaced symbolic definitions to prevent the model from memorizing specific symbols. The above sources are uniformly leveraged to capture the underlying connections between symbols from the data view.

From the framework aspect, we apply a two-stage continual tuning framework including the \emph{Injection Stage} and the \emph{Infusion Stage}. The \emph{Injection Stage} prioritizes the exploitation of the inherent connections between different symbols, thereby enabling the model to thoroughly learn a wide range of symbolic knowledge. After tuning LLaMA-2-Chat models with all collected symbolic data, we obtain \slmb variants. The \emph{Infusion Stage} focuses on balancing the model's dual capabilities by utilizing both symbolic data and general instruction tuning. After combining the general instruction-tuning data with the sampled symbolic data and tuning based on \slmb, we can obtain \slmi. Finally, Symbol-LLM series models are widely tested on both symbol-centric and NL-centric tasks, which are verified to exhibit substantial superiority.

\if
In short, 
We propose Symbol-LLM, a series of ``foundational'' and open-source models aiming at advancing the leverage of symbols to expand the effects of LLMs. 
The two aforementioned problems are addressed from both data and framework aspects. 
From the perspective of data, 
a collection of 34 text-to-symbol instruction tasks, covering  $\sim$20 different symbolic forms, are obtained from sampling off-the-shelf benchmarks, prompting GPT-4 or introducing symbol-evol strategies. 
They are supervised finetuned in a unified manner to capture symbol interrelations. From the framework aspect, this work designs a two-stage framework with \emph{Injection} and \emph{Infusion} stages. 
The former includes abundant text-to-symbol tasks to train on LLaMA-2-Chat models, obtaining \slmb series. The latter stage combines a collection of general instruction-tuning data with symbolic data. They are leveraged to continually train on \slmb models and obtain \slmi versions. Symbol-LLM series are widely tested on both symbol-centric and NL-centric tasks, which are verified to exhibit substantial superiority.
\fi

Our contributions are as follows, with key values and research insights attached in Appendix~\ref{key_contributions},~\ref{research_insights}:
\begin{itemize}[nosep,itemsep=1pt,leftmargin=0.1cm]
\item A comprehensive collection of text-to-symbol generation tasks is the first collection to treat symbolic data in a unified view and explore the underlying connections among symbols.
\item The open-sourced Symbol-LLM series models build a new foundation LLM with balanced symbolic and NL abilities.
\item Extensive experiments on both symbol- and NL-centric tasks are conducted to prove the superiority of \slm.
\end{itemize}

\section{Approach}

In this section, we first introduce the overall symbolic data collection procedure in Section~\ref{sec:data_collection} and then describe the two-stage tuning framework and the comprehensive test settings in Section~\ref{sec:framework}.

\subsection{Data Collection}
\label{sec:data_collection}

Conducting comprehensive symbolic knowledge injection and exploiting their interrelations requires a large collection of symbolic data.  However, achieving diverse knowledge coverage continues to be a significant hurdle in language modeling. Therefore, we curate an extensive collection of symbolic tasks, which is under-explored in NLP.

\begin{figure}[t]
\large
\centering
\includegraphics[scale=0.40]{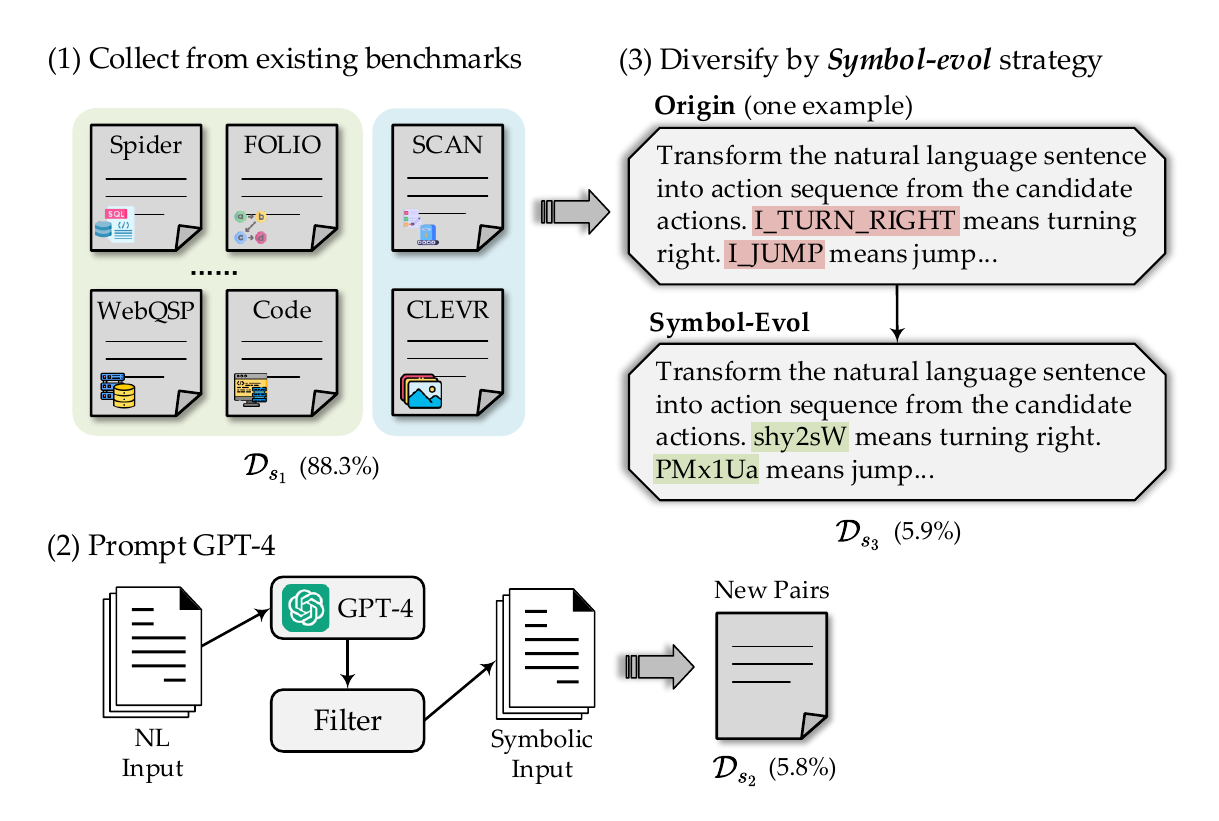}
\caption{Overview of the data collection procedure. It involves three key sources: (1) existing benchmarks, (2) new data generated via prompting GPT-4, and (3) new data synthesized using the \textit{Symbol-evol} strategy.}
\label{fig:symbol_evol}
\end{figure}

\begin{figure*}[t]
\large
\centering
\includegraphics[scale=0.23]{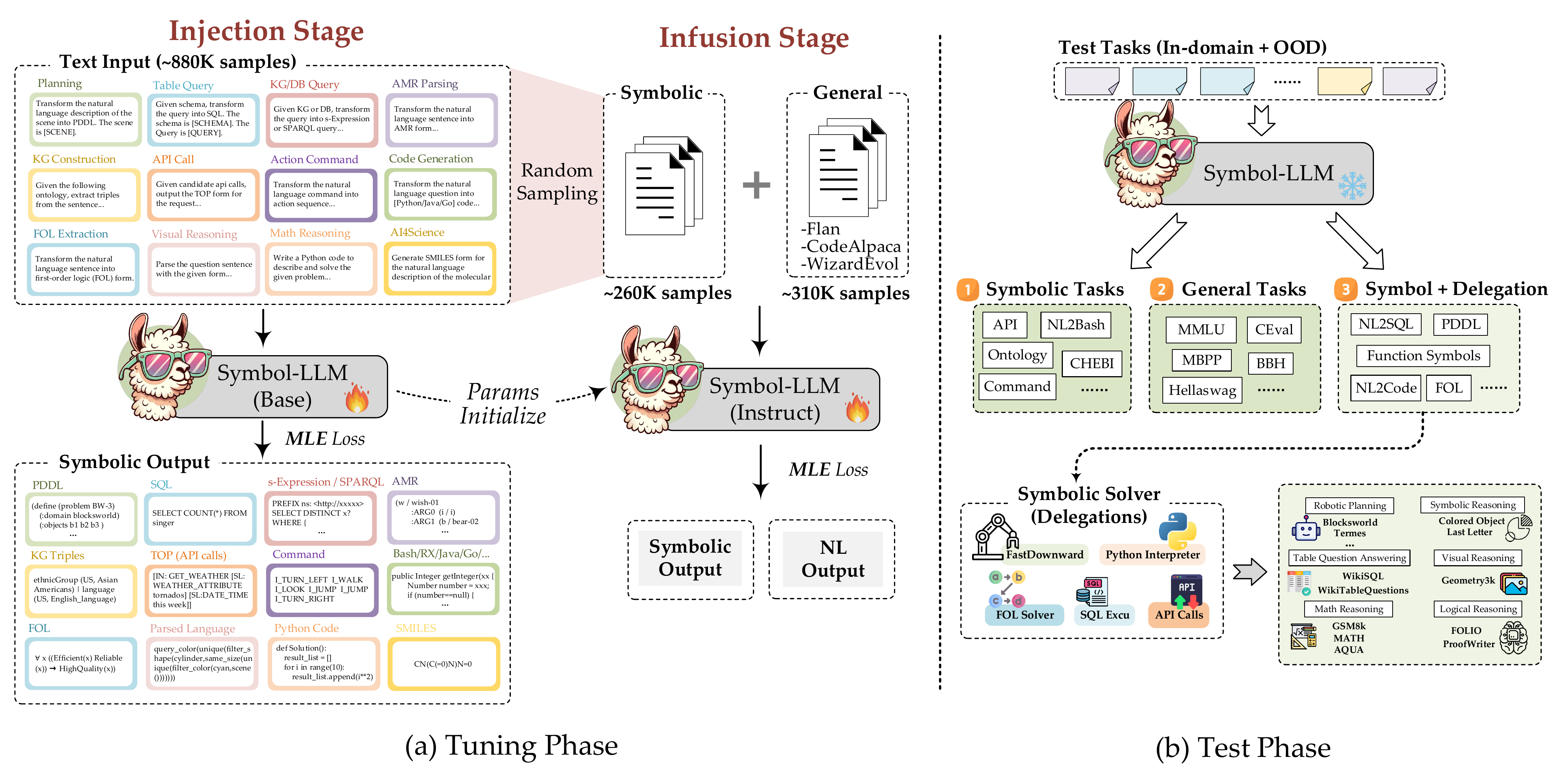}
\caption{Overall pipeline of Symbol-LLM. (a) is two-stage tuning framework, \emph{Injection} stage and \emph{Infusion} stage. (b) is the test phase with comprehensive settings, symbolic tasks, general tasks, and downstream tasks under the \emph{Symbol+Delegation} paradigm.}
\label{model}
\end{figure*}

The overview of the symbolic data collection procedure is shown in Figure~\ref{fig:symbol_evol}.  The ultimate symbolic dataset is $\mathcal{D}_{s} = \mathcal{D}_{s_1} \cup \mathcal{D}_{s_2} \cup \mathcal{D}_{s_3}$. Here,  $\mathcal{D}_{s_1}$ represents the existing benchmarks. The dataset $\mathcal{D}_{s_2}$ is a novel dataset, resulting from prompting GPT-4. $\mathcal{D}_{s_3}$ is another new dataset, generated by introducing the \textit{Symbol-evol} strategy. Generally, we compile a set of 34 text-to-symbol generation tasks, covering $\sim$20 different standard symbolic forms. To maintain the general capability in NL-centric tasks, this work also includes general instruction data $\mathcal{D}_g$. Details of each dataset are attached in Appendix~\ref{appendix:data_collection}.

\paragraph{$\mathcal{D}_{s_1}$: the existing symbolic datasets and benchmarks}

Previous efforts have been dedicated to specific symbolic forms, offering a natural and strong foundation for Symbol-LLM. We include plenty of text-to-symbol tasks from various data sources such as Spider~\cite{DBLP:conf/emnlp/YuZYYWLMLYRZR18}, MTOP~\cite{li-etal-2021-mtop}, SCAN~\cite{lake2018generalization}, and further shape them in the defined formats. Such collection is named as $\mathcal{D}_{s_1}$.



\paragraph{$\mathcal{D}_{s_2}$: novel text-to-symbol pairs by prompting GPT-4} 
While $\mathcal{D}_{s_1}$ has broad coverage,
it lacks certain text-to-symbol pairs in some crucial scenarios.
For example, 
some mathematical problems can be better handled when converted to programming language, 
but labeled samples are limited.
To address this,
we prompt GPT-4 to generate the corresponding symbolic outputs given the NL instructions, following~\citet{gao2023pal}.
Correct outputs judged by executing solvers (e.g., code interpreter) are retained to form new text-to-symbol pairs, 
constructing the collection $\mathcal{D}_{s_2}$.

\paragraph{$\mathcal{D}_{s_3}$: new samples generated by applying \emph{Symbol-evol} strategy} The above collection can cover a vast range of standard definitions of symbolic forms. However one concern is that large tuning data with the same symbolic definitions magnify LLM's propensity to memorize the patterns instead of truly learning to follow instructions. Thus, we introduce the \emph{Symbol-evol} strategy, expecting to enhance the diversity of symbolic systems.

The strategy of \emph{Symbol-evol}, as depicted in Figure~\ref{fig:symbol_evol}(3), is exemplified using \emph{SCAN} dataset~\cite{lake2018generalization}. In the original data collection, some action commands (in red background) are defined to control robots. Randomly generated strings (in green background) are leveraged to replace the original symbolic definitions. For example, the originally defined command \emph{I\_TURN\_RIGHT} is replaced by \emph{shY2sW}. In this way, diverse symbol instruction samples can be derived based on some original tasks in $\mathcal{D}_{s_1}$, forming the collection $\mathcal{D}_{s_3}$.

\paragraph{$\mathcal{D}_{g}$: general data} These collected data are from three sources: (i) sampled flan collection data~\cite{wei2021finetuned,longpre2023flan}; (ii) Code Alpaca instruction tuning data~\cite{codealpaca}; (iii) sampled Evol-data from WizardLM~\cite{xu2023wizardlm}. Full details are given in Appendix~\ref{appendix:general_collection}.

\subsection{Symbol-LLM}
\label{sec:framework}

\if
Considering that symbols and NL largely differ in format, it can easily lead to catastrophic forgetting of general knowledge if we purely overfit the symbolic knowledge. Meanwhile, the simple mixture of symbol- and NL-centric data for instruction tuning is sensitive to the mixing ratio and tends to obfuscate the model learning. Thus, we propose a two-stage tuning framework, including \emph{Injection} and \emph{Infusion} stage.

The whole approach is presented in Figure~\ref{model}, where we provide the overall pipeline for tuning and testing.
In the tuning part~(Fig.\ref{model}a), our approach is designed to address two problems above: (1) the symbolic interrelations; (2) the balance between symbol- and NL-centric capabilities, respectively from the data and framework perspectives.
In the test part~(Fig.\ref{model}b), we apply the tuned models to vast scenarios, including symbolic generation tasks, general tasks as well as the \emph{Symbol+Delegation} setting~\footnote{We refer to the paradigm that first generates symbolic representation and then delegate the solution to the external tools as the \emph{Symbol+Delegation} in this paper.}. 
\fi

The overview of Symbol-LLM is shown in Figure~\ref{model}, comprised of both the tuning and testing phases. 

The tuning framework, as illustrated in Fig.\ref{model}a, encompasses two stages: the \emph{Injection} stage and \emph{Infusion} stage. After the \emph{Injection} stage, we can obtain the \slmb model, which is expected to address various symbol-related scenarios. However, \emph{Injection} stage focuses on injecting symbolic knowledge into LLMs regardless of the general capability. 
But we also expect Symbol-LLM to maintain the necessary proficiency in general tasks, to achieve balanced symbol and NL interfaces for interaction and reasoning.
Thus, we introduce the \emph{Infusion} stage to obtain the \slmi. 

The test phase, represented in Fig.\ref{model}b, covers comprehensive settings on the symbolic and NL scenarios. 


\paragraph{Tuning Phase 1: Injection Stage}

In this stage, we purely focus on injecting various symbolic knowledge into LLMs by conducting supervised fine-tuning~(SFT) on the $\mathcal{D}_{s}$ collection. The training loss of \emph{Injection} stage is the maximum likelihood estimation~(MLE):
\begin{equation}
\small
    \mathcal{L}_{\mathrm{MLE}}(\mathcal{D}_{s}) = - \sum\limits_{i}{\mathrm{log}} \, p_{\theta}(y_i|s_i \oplus x_i),
\vspace{-0.2cm}
\end{equation}
where $p_{\theta}$ is the tunable LLM with parameters $\theta$, which is initialized from LLaMA-2-Chat models. $s_i \oplus x_i$ refers to the input format: the instruction~($s_i$) covering the task definition concatenates ($\oplus$) with the natural language query ($x_i$). And $y_i$ is the symbolic output.



\paragraph{Tuning Phase 2: Infusion Stage}
In this stage, we randomly sample $\mathcal{D}_{s}$ to obtain a subset $\mathcal{D}_{s'} \subset \mathcal{D}_{s}$,
the data are proportioned to ensure a fair distribution.
They are combined with general instruction tuning data $\mathcal{D}_{g}$ to form the training set in this stage. The loss function to be minimized is based on MLE:
\begin{equation}
\small
    \mathcal{L}_{\mathrm{MLE}}(\mathcal{D}_{s'} \cup \mathcal{D}_{g}) = - \sum\limits_{j}{\mathrm{log}} \, p_{\theta_1}(y_j|s_j \oplus x_j),
\vspace{-0.2cm}
\end{equation}
where the tunable parameters $\theta_1$ are initialized from \slmb. $s_j$, $x_j$, and $y_j$ are the instruction, input, and output for a single sample, respectively.


\paragraph{Testing Phase} This work presents comprehensive testing settings for border applications. For detailed task descriptions refer to Appendix~\ref{appendix:test}.

\begin{itemize}[nosep,itemsep=1pt,leftmargin=0.1cm]
\item Symbolic Tasks: Extensive symbolic generation tasks stress the unique advantages of addressing symbolic language beyond NL.
\item General Tasks: Classical benchmarks of general tasks are leveraged to verify the balanced capabilities in symbol- and NL-centric scenarios.
\item Symbol+Delegation Tasks: Verifying the effectiveness of LLM with symbolic-centric interface. We refer to this promising setting as \emph{Symbol+Delegation}, where the model first generates the symbolic representation of the question and then relies on the external solvers for solution~(e.g., Python interpreter, SQL execution).
\end{itemize}






\section{Experiments}
In this section, we fully evaluate Symbol-LLM\footnote{Unless otherwise specified, Symbol-LLM represents the final model after two stages (i.e., \emph{Instruct} version).} on three parts of experiments: the symbolic tasks in Sec.~\ref{exp:symbolic_tasks}, the general tasks in Sec.~\ref{exp:general_tasks}, and the Symbol+Delegation tasks in Sec.~\ref{exp:symbol_delegation}. The implementation details refer to Appendix~\ref{appendix:test} and Appendix~\ref{appendix:exp_settings}. The overall performances of Symbol-LLM are concluded in Appendix~\ref{appendix:overall_performances}.

\subsection{Symbolic Tasks}
\label{exp:symbolic_tasks}

\begin{table*}[t]
\centering
\footnotesize
\resizebox{\linewidth}{!}{
\begin{tabular}{cl|c|cc|ccc|ccc}
    \toprule
    \multicolumn{2}{c|}{\multirow{2}{*}{\textbf{Domains / Tasks}}} &\multirow{2}{*}{\textbf{Metrics}} & \multicolumn{2}{c|}{\textbf{Close-Source}}  & \multicolumn{3}{c|}{\textbf{Open-source (7B)}} & \multicolumn{3}{c}{\textbf{Open-source (13B)}} \\
    & & &\makecell[c]{\scriptsize{GPT-3.5}} &\makecell[c]{\scriptsize{Claude-1}} &\makecell[c]{\scriptsize{LLaMA-2-Chat}} &\makecell[c]{\scriptsize{Single SFT}} &\makecell[c]{\scriptsize{Symbol-LLM}} &\makecell[c]{\scriptsize{LLaMA-2-Chat}} &\makecell[c]{\scriptsize{Single SFT}} &\makecell[c]{\scriptsize{Symbol-LLM}} \\
    \midrule
    \multirow{4}{*}{Planning} & Blocksworld &BLEU &96.54 &91.35 &85.16 &97.40 &\textbf{99.02}\textsubscript{0.11} &31.27 &97.06 &\textbf{99.02}\textsubscript{0.12} \\
    & Termes &BLEU &74.73 &26.94 &53.08 &\textbf{67.46} &48.69\textsubscript{4.48} &59.30 &68.63 &\textbf{90.09}\textsubscript{0.66} \\
    & Floortile &BLEU &54.23 &13.94 &59.41 &78.07 &\textbf{95.84}\textsubscript{0.45} &0.00 &74.22 &\textbf{95.24}\textsubscript{0.21} \\
    & Grippers &BLEU &99.90 &90.91 &86.15 &94.84 &\textbf{98.53}\textsubscript{0.51} &95.36 &97.46 &\textbf{98.89}\textsubscript{0.32}  \\
    \midrule
    \multirow{3}{*}{SQL} & Spider &EM &42.60 &32.70 &16.50 &\textbf{65.30} &63.80\textsubscript{0.09} &10.30 &68.20 &\textbf{69.20}\textsubscript{0.05}  \\
    & Sparc &EM &29.90 &28.60 &12.50 &\textbf{55.40} &55.00\textsubscript{0.07} &10.20 &57.50 &\textbf{58.90}\textsubscript{0.05} \\
    & Cosql &EM &18.80 &22.70 &9.30 &\textbf{51.30} &48.20\textsubscript{0.07} &1.20 &\textbf{54.60} &52.70\textsubscript{0.07} \\
    \midrule
    \multirow{3}{*}{KG / DB} & WebQSP &F1 &36.49 &41.37 &0.09 &\textbf{84.93} &84.43\textsubscript{0.26} &0.00 &84.80 &\textbf{85.29}\textsubscript{0.23} \\
    & GrailQA &EM &28.52 &25.56 &0.00 &\textbf{80.58} &79.24\textsubscript{0.21} &0.06 &\textbf{81.82} &81.17\textsubscript{0.22}  \\
    & CompWebQ &EM &0.00 &0.00 &0.00 &\textbf{56.30} &50.98\textsubscript{0.46} &0.00 &\textbf{59.02} &54.94\textsubscript{0.52} \\
    \midrule
    \multirow{3}{*}{AMR} &AMR3.0 &Smatch &18.00 &10.00 &6.00 &\textbf{55.00} &54.00\textsubscript{0.00} &2.00 &55.00 &55.00\textsubscript{0.00}  \\
    &AMR2.0 &Smatch &14.00 &12.00 &7.00 &\textbf{46.00} &45.00\textsubscript{0.00} &1.00 &\textbf{47.00} &46.00\textsubscript{0.00}\\
    &BioAMR &Smatch &23.00 &3.00 &24.00 &\textbf{80.00} &78.00\textsubscript{0.22} &0.00 &80.00 &80.00\textsubscript{0.00} \\
    \midrule
    \multirow{2}{*}{Ontology} &Tekgen &F1 &8.92 &1.86 &4.50 &56.69 &\textbf{57.34}\textsubscript{0.02} &6.24 &58.49 &\textbf{58.55}\textsubscript{0.13} \\
    &Webnlg &F1 &28.34 &8.89 &7.38 &\textbf{63.75} &60.42\textsubscript{0.05} &17.23 &62.13 &\textbf{63.08}\textsubscript{0.08}  \\
    \midrule
    \multirow{3}{*}{API} & MTOP &EM &3.80 &8.40 &0.00 &\textbf{84.80} &84.40\textsubscript{0.15} &0.00 &86.20 &\textbf{86.60}\textsubscript{0.12} \\
    & TOPv2 &EM &6.60 &7.60 &0.00 &\textbf{86.60} &85.80\textsubscript{0.06} &0.00 &\textbf{87.20} &85.20\textsubscript{0.06}  \\
    & NLmaps &EM &30.88 &16.77 &2.00 &91.95 &\textbf{92.18}\textsubscript{0.03} &3.60 &\textbf{92.38} &92.21\textsubscript{0.02} \\
    \midrule
    \multirow{1}{*}{Command} & SCAN &EM &15.09 &15.97 &0.00 &98.23 &\textbf{98.35}\textsubscript{0.00} &0.00 &98.99 &\textbf{99.28}\textsubscript{0.00} \\
    \midrule
    \multirow{5}{*}{Code} & NL2BASH &BLEU &54.19 &42.24 &23.29 &59.22 &\textbf{60.25}\textsubscript{0.18} &19.06 &60.68 &\textbf{60.76}\textsubscript{0.12} \\
    &NL2RX &BLEU &38.60 &18.30 &5.91 &\textbf{85.25} &85.08\textsubscript{0.12} &0.00 &\textbf{85.55} &84.97\textsubscript{0.28}  \\
    &NL2Python &BLEU &37.01 &36.73 &26.68 &38.19 &\textbf{39.79}\textsubscript{0.28} &34.94 &40.35 &\textbf{40.76}\textsubscript{0.32}  \\
    &NL2Java &BLEU &24.88 &22.79 &25.77 &27.33 &\textbf{28.08}\textsubscript{0.22} &23.49 &\textbf{28.47} &28.25\textsubscript{0.19}  \\
    &NL2Go &BLEU &19.08 &26.65 &24.00 &\textbf{30.77} &29.19\textsubscript{0.39} &1.26 &24.75 &\textbf{30.31}\textsubscript{0.33} \\
    \midrule
    \multirow{3}{*}{FOL} & FOLIO &LE &60.65 &53.47 &33.98 &\textbf{90.81} &90.58\textsubscript{0.01} &28.79 &\textbf{91.59} &90.65\textsubscript{0.02} \\
    & MALLS &LE &69.15 &30.46 &55.13 &\textbf{89.24} &88.88\textsubscript{0.03} &11.71 &89.41 &\textbf{89.50}\textsubscript{0.01} \\
    & LogicNLI &LE &73.11 &69.16 &39.95 &\textbf{100.00} &99.97\textsubscript{0.00} &32.26 &99.99 &\textbf{100.00}\textsubscript{0.00}  \\
    \midrule
    \multirow{3}{*}{Visual} &GQA &EM &7.55 &7.70 &0.30 &\textbf{85.65} &85.50\textsubscript{0.01} &8.85 &\textbf{86.10} &85.95\textsubscript{0.01} \\
    & CLEVR &EM &6.35 &5.90 &0.25 &86.35 &\textbf{94.80}\textsubscript{0.11}&1.15 &92.20 &\textbf{95.60}\textsubscript{0.09} \\
    & Geometry3k &EM &65.25 &40.84 &36.88 &93.92 &\textbf{95.13}\textsubscript{0.02} &52.17 &94.52 &\textbf{95.67}\textsubscript{0.03} \\
    \midrule
    \multirow{3}{*}{Math} & GSM8K-Code &BLEU &82.20 &63.42 &53.66 &\textbf{85.31} &84.14\textsubscript{0.32} &72.29 &84.01 &\textbf{84.42}\textsubscript{0.23} \\
    & AQUA-Code &BLEU &67.48 &48.88 &39.25 &66.27 &\textbf{67.05}\textsubscript{0.62} &55.13 &65.66 &\textbf{67.20}\textsubscript{0.77}  \\
    & MATH-Code &BLEU &56.48 &48.87 &29.88 &56.43 &\textbf{57.36}\textsubscript{0.89} &48.85 &\textbf{58.24} &56.97\textsubscript{1.12} \\
    \midrule
    \multirow{1}{*}{AI4Science} &CheBi-20 &EM &1.15 &0.30 &0.00 &40.36 &\textbf{58.97}\textsubscript{0.92} &0.00 &46.82 &\textbf{65.27}\textsubscript{0.89} \\
    \midrule
    \rowcolor{gray!20} \multicolumn{3}{c|}{\multirow{1}{*}{\textbf{Average Performance}}} &32.27 &25.04 &22.59 &71.46 &\textbf{71.88} &18.46 &72.32 &\textbf{74.34} \\
    \bottomrule
\end{tabular}}
\caption{Main results on 34 symbolic tasks. Better results with the same model size are marked in bold. \emph{GPT-3.5}, \emph{Claude-1}, and \emph{LLaMA-2-Chat} column presents the baseline performances of prompting these models under the few-shot setting. \emph{Single-SFT} represents the models fine-tuned with single-domain samples based on LLaMA-2-Chat. \emph{Symbol-LLM} column represents the final obtained model after two-stage tuning, with beam size 1 for decoding. The subscript of the result denotes the variances between three generations (with different beam sizes).}
\vspace{-0.2cm}
\label{text2symbol}
\end{table*}

\begin{table*}[t]
\centering
\footnotesize
\begin{tabular}{l|ccccc|c}
    \toprule
    \multicolumn{1}{c|}{\multirow{2}{*}{\textbf{Models}}} &\multicolumn{5}{c|}{\textbf{MMLU (5-shot)}} &\multicolumn{1}{c}{\textbf{BBH (0-shot)}} \\
    &Humanities &SocialSciences &STEM &Others &Average &Average  \\
    \hline
    \multicolumn{7}{c}{\cellcolor{gray!20} Close-source LLMs} \\
    \hline
    GPT-3.5 &54.90 &69.58 &49.73 &66.75 &59.74 &56.84 \\
    Claude-1 &56.60 &74.15 &53.66 &60.35 &62.09 &47.01 \\
    \hline
    \multicolumn{7}{c}{\cellcolor{gray!20} Open-source LLMs (7B)}
    \\
    \hline
    LLaMA-2-Chat &\underline{42.47} &\underline{52.49} &\underline{36.94} &\underline{52.47} &\underline{45.78} &35.01  \\
    CodeLLaMA-Instruct &39.47 &46.31 &35.95	&45.34 &41.57 &\underline{35.69} \\
    \textbf{\slmb} &40.04 &46.28 &33.73 &47.16 &41.70 &33.82 \\
    \textbf{\slmi} &\textbf{46.33} &\textbf{57.20} &\textbf{40.39} &\textbf{54.53} &\textbf{49.30} &\textbf{39.30} \\
    \hline
    \multicolumn{7}{c}{\cellcolor{gray!20} Open-source LLMs (13B)}
    \\
    \hline
    LLaMA-2-Chat &\textbf{49.52} &\textbf{62.43} &\textbf{43.84} &\textbf{60.02} &\textbf{53.55} &\underline{36.99} \\
    CodeLLaMA-Instruct &33.88 &41.92 &34.69	&42.17 &37.73 &36.71  \\
    \textbf{\slmb} &45.67 &55.67 &40.09	&53.89 &48.56 &35.26 \\
    \textbf{\slmi} &\underline{48.88} &\underline{62.14} &\underline{43.44}	&\underline{57.93} &\underline{52.71} &\textbf{44.09} \\
    \bottomrule
\end{tabular}
\caption{Results on general tasks. We include 57 tasks in the MMLU benchmark for testing under the 5-shot setting~\cite{hendryckstest2021}, while we select 21 tasks in BBH under the 0-shot setting following~\citet{eval-harness}. Best results are marked in bold while sub-optimal results are underlined (same for the following tables).}
\label{general}
\end{table*}


Table \ref{text2symbol} presents the results of 34 symbolic generation tasks.
For model comparison, we include GPT-3.5, Claude-1, LLaMA-2-Chat, and the optimized model after single-domain SFT on LLaMA-2-Chat.
Limited by space, we compare other baseline models (e.g., CodeLLaMA-Instruct) in Appendix~\ref{appendix:overall_performances}. For extra OOD symbolic tasks, please refer to Appendix~\ref{symbol_ood}. The main findings are as follows:

\paragraph{Symbol-LLM largely enhances the symbol-related capabilities of LLM.}
In comparison with the LLaMA-2-Chat model, Symbol-LLM presents overwhelming advantages in symbolic tasks. It improves the baseline performances of 7B and 13B by 49.29\% and 55.88\%, respectively. Also, cutting-edge close-source LLMs like GPT-3.5 and Claude-1, are far behind Symbol-LLM, with the minimum gaps of 39.61\% (GPT-3.5 v.s. Symbol-LLM-7B). In short, Symbol-LLM brings huge advantages in symbolic scenarios.

\paragraph{The unified modeling helps Symbol-LLM successfully capture the intrinsic relationships between different symbols.}
Fine-tuning LLaMA-2-Chat on single-domain tasks fully overfits domain-specific symbolic forms, as shown in \textit{Single SFT} of Table~\ref{text2symbol}. Compared with it, Symbol-LLM shows better performances, with averaged 0.42\% and 2.02\% gains in 7B and 13B. It verifies that the unified modeling of various symbolic forms is beneficial to capturing symbolic interrelations.

\subsection{General Tasks}
\label{exp:general_tasks}

To verify Symbol-LLM's power in tackling NL-centric tasks, we conduct the experiments on two widely-used benchmarks, MMLU and BIG-Bench-Hard (BBH). Results are shown in Table \ref{general}.

\paragraph{Competitive performances in general tasks are maintained in Symbol-LLM.}
Overall, \slm is well optimized with the two-stage framework in keeping general abilities. For 7B models, \slmi shows consistent superiority on MMLU and BBH benchmarks, with $\sim$4\% gains compared with LLaMA-2-Chat. For 13B models, although \slmi slightly falls behind its LLaMA counterpart, it achieves 7.20\% performance advantages in BBH. The superiority on average is obvious. While Symbol-LLM may not yet match the performance of closed-source LLMs, its well-rounded general capability is notable. 

Notably, a broader scope of evaluation on general tasks is attached in Appendix~\ref{appendix:exp_general_tasks}.

\subsection{Symbol+Delegation Tasks}
\label{exp:symbol_delegation}
A wide range of experiments are done under the \emph{Symbol+Delegation} paradigm, covering the fields of math reasoning, symbolic reasoning, logical reasoning, robotic planning, visual reasoning as well as table question answering.
For detailed settings, please refer to Appendix~\ref{appendix:symbol+delegation}.
Limited by space, we only present the results of the math reasoning in the main paper. The remaining parts are attached in Appendix~\ref{appendix:exp_symbol_delegation}.

\begin{table*}[t]
\centering
\footnotesize
\resizebox{\linewidth}{!}{
\begin{tabular}{lc|ccccccccc}
    \toprule
    \textbf{Models} & \textbf{Del.} &\textbf{GSM8k} & \textbf{MATH} & \textbf{GSM-Hard} &\textbf{SVAMP} &\textbf{ASDiv} &\textbf{ADDSUB} &\textbf{SingleEQ} &\textbf{SingleOP} &\textbf{MultiArith}\\
    \hline
    \multicolumn{2}{l|}{\textbf{Is OOD Setting}} &\color{red}\usym{2717}&\color{red}\usym{2717} &\color{ao(english)}\checkmark &\color{ao(english)}\checkmark &\color{ao(english)}\checkmark &\color{ao(english)}\checkmark &\color{ao(english)}\checkmark &\color{ao(english)}\checkmark &\color{ao(english)}\checkmark\\
    \hline
    \multicolumn{11}{c}{\cellcolor{gray!20} Close-source LLMs} \\
    \hline
    GPT-3.5 &\color{ao(english)}\checkmark &4.60 &1.05 &4.62 &5.10 &6.30 &1.01 &3.94 &8.54 &17.33 \\
    GPT-3.5 (3-shot) &\color{ao(english)}\checkmark &76.04 &36.80 &62.09 &83.40 &85.73 &87.59 &96.46 &90.74 &96.67 \\
    Claude-1 &\color{ao(english)}\checkmark &11.14 &1.07 &9.02 &10.30 &6.30 &5.06 &4.53 &0.36 &12.67 \\
    Claude-1 (3-shot) &\color{ao(english)}\checkmark &58.07 &13.17 &43.75 &78.90 &74.19 &79.49 &88.19 &87.72 &91.83 \\
    \hline
    \multicolumn{11}{c}{\cellcolor{gray!20} Open-source LLMs (7B)} \\
    \hline
    LLaMA-2-Chat (3-shot) &\color{ao(english)}\checkmark &12.21 &1.32 &10.69 &22.00 &25.86 &29.11 &27.36 &39.15 &23.17 \\
    CodeLLaMA-Instruct (3-shot)$\dagger$ & \color{ao(english)}\checkmark &34.00 &16.60 &33.60 &59.00 &61.40 &\multicolumn{4}{c}{Average performance  79.60}	\\
    WizardMath$\dagger$ &\color{red}\usym{2717} &54.90 &10.70 &- &57.30 &- &- &- &- &- \\
    MAmmoTH$\dagger$ &\color{ao(english)}\checkmark &51.70 &\textbf{31.20} &- &66.70 &- &- &- &- &- \\
    \textbf{\slmb} &\color{ao(english)}\checkmark &\textbf{61.14} &\underline{28.24} &\textbf{52.62} &\underline{72.50} &\textbf{78.34} &\textbf{89.62} &\textbf{97.83} &\textbf{96.26} &\textbf{99.67} \\
    \textbf{\slmi} &\color{ao(english)}\checkmark &\underline{59.36} &26.54	&\underline{48.98} &\textbf{72.80} &\underline{75.76} &\underline{87.85} &\underline{96.26} &\underline{93.24} &\underline{99.00} \\
    \hline
    \multicolumn{11}{c}{\cellcolor{gray!20} Open-source LLMs (13B)} \\
    \hline
    LLaMA-2-Chat (3-shot) &\color{ao(english)}\checkmark &34.87 &6.07 &28.96 &45.00 &46.61 &45.57 &47.05 &56.76 &56.67 \\
    CodeLLaMA-Instruct (3-shot)$\dagger$ &\color{ao(english)}\checkmark &39.90 &19.90 &39.00 &62.40 &65.30 &\multicolumn{4}{c}{Average performance  86.00} \\
    WizardMath$\dagger$ &\color{red}\usym{2717} &63.90 &14.00 &- &64.30 &- &- &- &- &- \\
    MAmmoTH$\dagger$ &\color{ao(english)}\checkmark &61.70 &\textbf{36.00} &- &72.40 &- &- &- &- &- \\
    \textbf{\slmb} &\color{ao(english)}\checkmark &\textbf{68.69} &\underline{33.39} &\textbf{58.53} &\textbf{78.80} &\textbf{80.15} &\underline{91.14} &\textbf{96.85} &\textbf{95.55} &\underline{98.83} \\
    \textbf{\slmi} &\color{ao(english)}\checkmark &\underline{65.58} &31.32	&\underline{55.57} &\underline{76.80} &\underline{79.01} &\textbf{91.90} &\textbf{96.85} &\underline{94.84} &\textbf{99.33} \\
    \bottomrule
\end{tabular}}
\caption{Results on Math Reasoning. Del. represents whether uses delegation (i.e., Python Interpreter for math reasoning tasks). Results are under the zero-shot setting unless otherwise stated (the following tables share the same setting). $\dagger$ indicates that the results are reported from~\citet{luo2023wizardmath}, \citet{yue2023mammoth} and \citet{gou2023tora}. 
}
\label{math}
\end{table*}

We select 9 commonly used math datasets for testing, including both in-domain and OOD tasks. To demonstrate the surprising performances of Symbol-LLM, we also include several math-domain LLMs (e.g., WizardMath~\cite{luo2023wizardmath}, MAmmoTH~\cite{yue2023mammoth}) as strong baselines. Comparison results are presented in Table~\ref{math}.

\paragraph{Advanced abilities in math reasoning are possessed by Symbol-LLM.}
GSM8K and MATH are widely used to evaluate the math reasoning capabilities of LLMs. Compared with recent math-domain LLMs, Symbol-LLM presents great competitive results on them. Especially on GSM8K, Symbol-LLM consistently wins all strong baselines with great margins with all the model variants. On the MATH dataset, Symbol-LLM merely falls behind MAmmoTH, which is a strong LLM specially designed for math reasoning tasks. Notably, MAmmoTH includes GSM8K and MATH in the tuning stage and it also uses delegation (i.e., Python Interpreter) for inference, thus our comparisons are fair. Similar superiority is also observed under the OOD tasks (e.g., SVAMP).

\paragraph{Symbol-LLM exhibits competitive performances in extrapolating to OOD tasks.}
More surprisingly, Symbol-LLM consistently presents its significant superiority among all 7 OOD math tasks.
Even compared with GPT-3.5 under the three-shot setting, our Symbol-LLM-7B series won 4 (out of 7) OOD tasks under the zero-shot setting. 
As we scale the model size to 13B, obvious performance improvements are observed in most of the tasks.
These findings verify the prospects of \slm under the Symbol+Delegation paradigm.

\section{Analysis}
In this section, we include the ablation studies~(Sec.\ref{ablation_studies}) and the analysis on \emph{Alignment} and \emph{Uniformity}~(Sec.\ref{align_and_uniform}). Notably, additional supplementary experiments are attached in Appendix~\ref{appendix:supplementary_exp}, including tuning design choices~\ref{appendix:tuning_design_choice}, comparison with single SFT~\ref{appendix:single_sft}, and new symbol extrapolation~\ref{appendix:extrapolation}.

\subsection{Ablation Studies}
\label{ablation_studies}

Here we present two ablation experiments from both the framework and data views: (1) tuning only in one stage, and (2) tuning only on general data collection. For a fair comparison, we introduce two settings for one-stage tuning. The first setting (named \emph{One-stage 1.46M}) simply mixes $\mathcal{D}_{s}$, $\mathcal{D}_{s'}$ and $\mathcal{D}_{g}$, regardless of sample overlap. The second setting (named \emph{One-stage 1.20M}) mixes $\mathcal{D}_{s}$ and $\mathcal{D}_{g}$, which ensures consistency in diversity and avoids duplication. The model exclusively fine-tuned on general task $\mathcal{D}_{g}$ is referred to as \emph{General-only}. Comparison results are shown in Table~\ref{exp:comparison}. 

\begin{table*}[t]
\centering
\footnotesize
\begin{tabular}{l|cccc|cccc}
    \toprule
    \multicolumn{1}{c|}{\multirow{2}{*}{\textbf{Models}}} &\multicolumn{4}{c|}{\textbf{7B Models}} &\multicolumn{4}{c}{\textbf{13B Models}} \\
     &\textbf{Symbolic} &\textbf{General} &\textbf{Symbol+Del.} &\textbf{Avg.} &\textbf{Symbolic} &\textbf{General} &\textbf{Symbol+Del.} &\textbf{Avg.}  \\
    \hline
    \slm &71.88 &44.30 &52.54 &56.24 &74.34 &48.40 &60.45 &61.06 \\
    \hline
    One-stage 1.20M &70.38 &45.24 &47.27 &54.30 &70.59 &48.29 &53.99 &57.62 \\
    \quad\quad\quad $\Delta$ &{\color{red} (+1.50)} &{\color{ForestGreen} (-0.94)} &{\color{red} (+5.27)} &{\color{red} (+1.94)} &{\color{red} (+3.75)} &{\color{red} (+0.11)} &{\color{red} (+6.46)} &{\color{red} (+3.44)} \\
    One-stage 1.46M &72.75 &44.44 &49.31 &55.50 &73.71 &46.59 &52.67 &57.66 \\
    \quad\quad\quad $\Delta$ &{\color{ForestGreen} (-0.87)} &{\color{ForestGreen} (-0.14)} &{\color{red} (+3.13)} &{\color{red} (+0.74)} &{\color{red} (+0.63)} &{\color{red} (+1.81)} &{\color{red} (+7.78)} &{\color{red} (+3.40)} \\
    General-only &28.66	&46.21 &28.17 &34.35 & 31.35 &49.72 &31.49 &37.52 \\
    \quad\quad\quad $\Delta$ &{\color{red} (+43.22)} &{\color{ForestGreen} (-1.91)} &{\color{red} (+24.37)} &{\color{red} (+11.89)} &{\color{red} (+42.99)} &{\color{ForestGreen} (-1.32)} &{\color{red} (+28.96)} &{\color{red} (+23.54)} \\
    \bottomrule
\end{tabular}
\caption{Comparison experiments. \emph{Avg.} denotes the simple averaged performances on the symbolic tasks, general tasks, and Symbol+Delegation tasks.}
\vspace{-0.2cm}
\label{exp:comparison}
\end{table*}

\paragraph{Two-stage tuning framework shows superiority over one-stage, especially for 13B.} Simply mixing the training data in one stage is prone to affecting the symbol-related tasks. Especially under the \emph{Symbol+Delegation} setting, the two-stage framework witnesses 3$\sim$6$\%$ advantages over the one-stage models. In the 13B model comparison, our two-stage framework consistently demonstrates superiority across symbolic tasks, general tasks, and \emph{Symbol+Delegation} tasks.

\paragraph{The incorporation of symbolic data yields a modest impact on the performances of general tasks.}
Compared with \emph{General-only}, \slmi is optimized to largely enhance the symbol-centric capabilities. Meanwhile, it maintains the capability to address general NL-centric tasks without significant sacrifices (< 2\%).

\subsection{Alignment and Uniformity}
\label{align_and_uniform}

Motivated by~\cite{wang2020understanding,gao2021simcse}, we include \emph{Alignment} and \emph{Uniformity} metrics to delve into the factors contributing to the superiority of Symbol-LLM.

\begin{figure}[t]
\large
\centering
\includegraphics[scale=0.58]{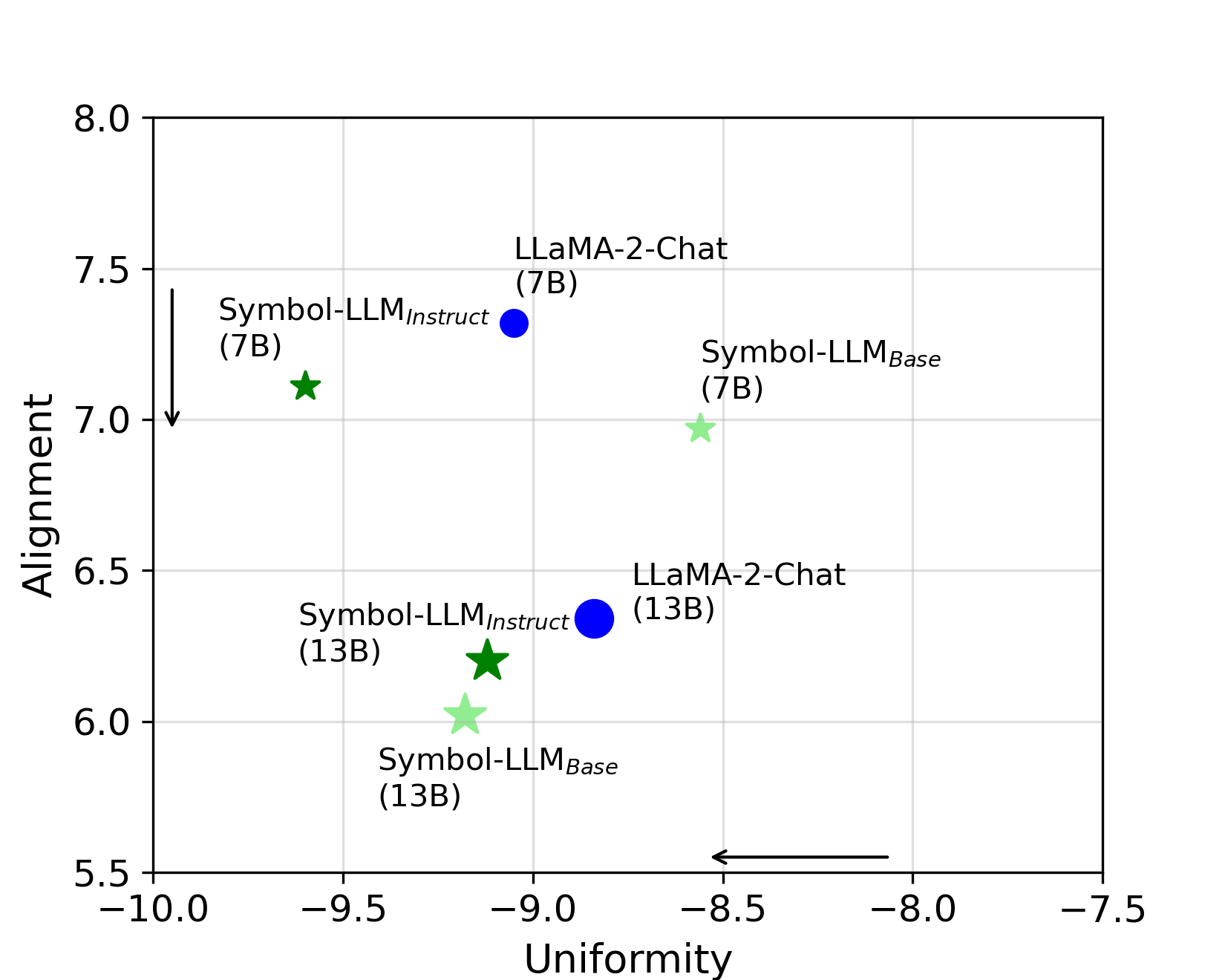}
\caption{Visualization of \emph{Alignment-Uniformity}. Both metrics are inversely related, which means a lower value indicates better performance.}
\vspace{-0.2cm}
\label{main_align_uniform}
\end{figure}

\emph{Alignment} measures the representation similarity within each symbolic form, while \emph{Uniformity} quantifies the uniformity of all the symbolic representations. They are based on Eq.~\ref{eq:cal_align} and Eq.~\ref{eq:cal_uniform} respectively in Appendix~\ref{appendix:align_uniform}. The calculation results are visualized in Figure~\ref{main_align_uniform}. Further, we extend the definition to measure the interrelations between any two symbolic forms, based on Eq.~\ref{eq:cal_align_2}. Limited by space, we only include a part of the symbolic forms for illustration and present the results of 13B models in Figure~\ref{main_alignment} in Appendix~\ref{appendix:align_uniform}.

The item-wise conclusions are listed as follows:

\noindent \textbf{Symbol-LLM optimizes symbol distinctiveness and overall expressiveness in the embedding space.}
From Fig.~\ref{main_align_uniform}, compared with the LLaMA-2-Chat models, \slm series is optimized towards superior \emph{Alignment} and \emph{Uniformity}. It ensures the discernment of shared features within each symbolic form, simultaneously enhancing the overall information entropy. 
Specifically for the 7B model, the two-stage framework effectively maintains a balance of uniformity, preventing the collapse of the embedding space.

\noindent \textbf{Symbol-LLM excels at capturing symbolic interrelations.}
From Fig.~\ref{main_alignment}, the LLaMA-2-Chat model exhibits significant representation sparsity between symbolic forms. Even under the same form (e.g., \emph{Bash}, \emph{FOL}), the features are scattered. On the contrary, \slm largely enhances the perception of symbolic interrelations by (1) achieving better alignments between symbols (e.g., \emph{Python-AMR} and \emph{CheBi-RX}) and (2) pulling closer sample features within each symbolic form (e.g., FOL).

\begin{figure}[t]
    \centering
    \begin{minipage}[t]{0.49\linewidth}
        \large
        \centering
        \includegraphics[scale=0.27]{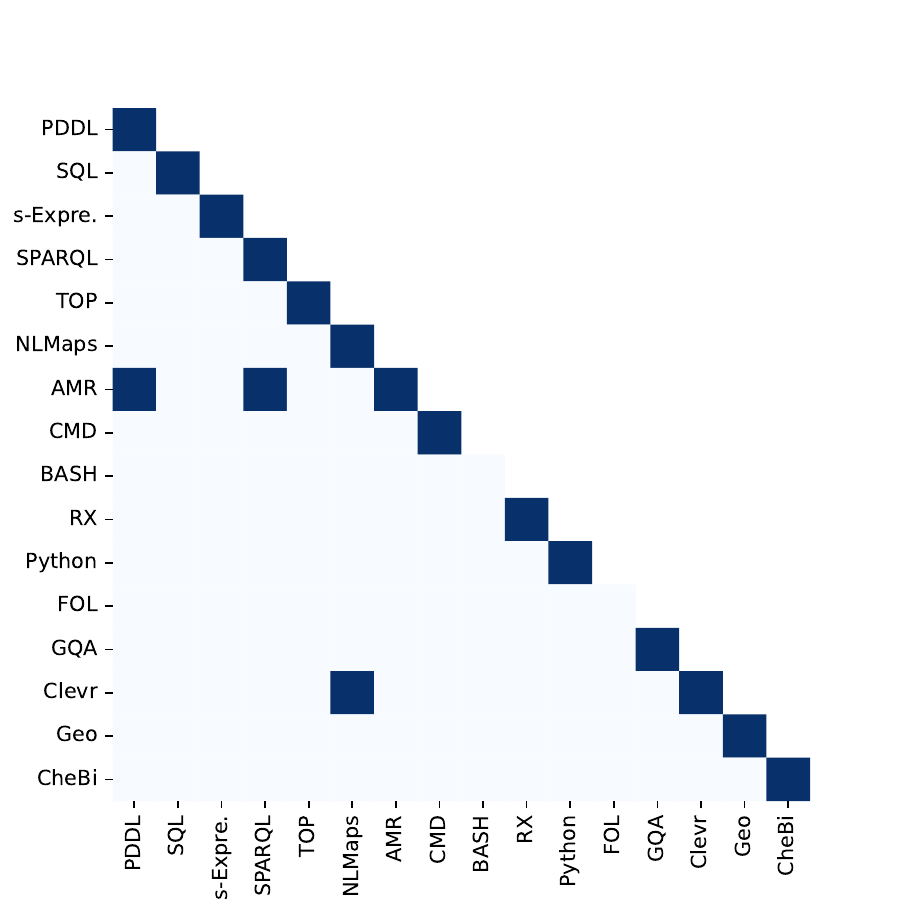}
        \subcaption{LLaMA-2-Chat (13B)}
        \label{align_llama_13b}
    \end{minipage}
    \begin{minipage}[t]{0.49\linewidth}
        \large
        \centering
        \includegraphics[scale=0.27]{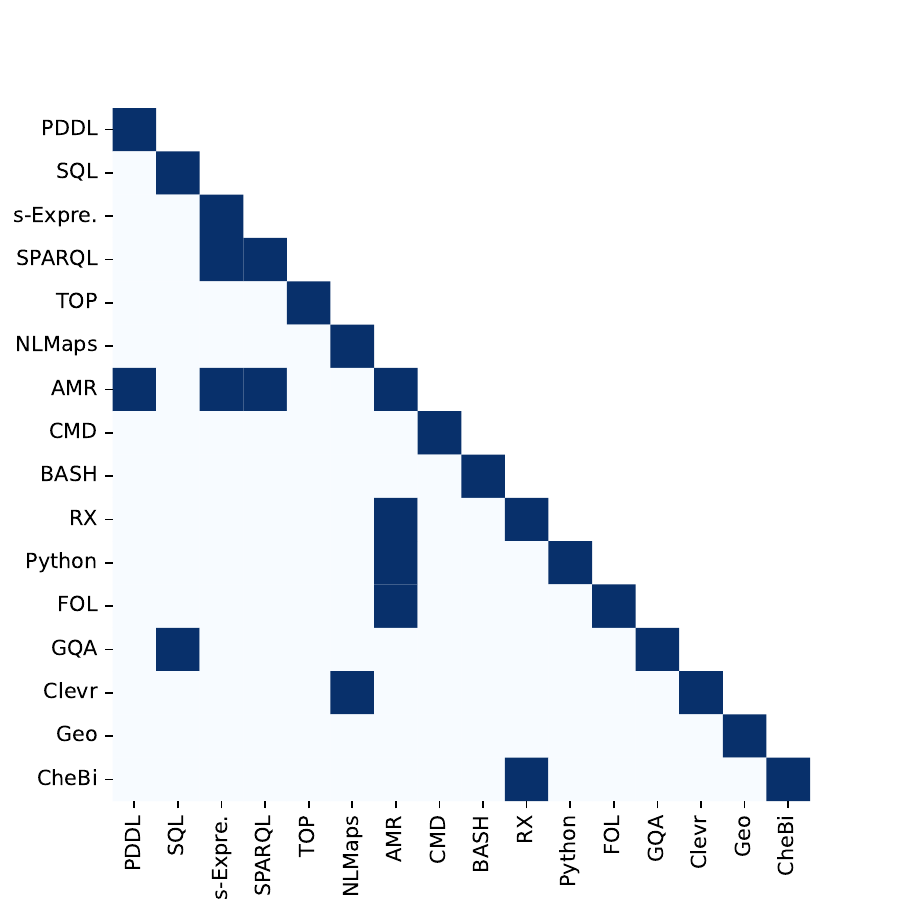}
        \subcaption{\slm (13B)}
        \label{align_symbolllm_13b}
    \end{minipage}
    \caption{Visualization of the alignment relations between symbols after binarization. Dark blue denotes a close relation between two symbols in the representation. Limited by space, we only showcase 13B models. More illustrations refer to Appendix~\ref{appendix:align_uniform}.}
    \label{main_alignment}
\end{figure}

\section{Related Works}
\vspace{-0.2cm}
\paragraph{Large Language Models}
Plenty of recent efforts have been made to develop foundation language models~\cite{zhao2023survey}, 
which are expected to promote the subsequent applications, 
such as AI agents~\cite{wang2023survey,sun2023corex,wu2024copilot,cheng2024seeclick}. 
These works on LLMs are universally categorized into closed-source and open-source models. 
Close-source LLMs, represented by GPT-4~\cite{OpenAI2023GPT4TR}, Claude, PaLM~\cite{chowdhery2022palm}, have greatly shaped our daily life through NL-centric interactions. 
However, 
their closed-source and black-box properties limit further optimization. 
Under such circumstances, open-source LLMs~\cite{zeng2022glm,jiang2023mistral,touvron2023llama2} receive significant attention because of their tunable and small-scale properties. 
However, 
current attempts on these LLMs mainly explore NL-centric abilities, 
which treats NL as the interface to express knowledge and achieve interactive reasoning. 
In contrast,
our work focuses on improving the symbol-centric capabilities of open-source LLM, 
which leads to a balanced symbol-centric and NL-centric foundational LLM.

\paragraph{Instruction Tuning}
To make LLMs capable of following human instructions, 
instruction fine-tuning~\cite{zhang2023instruction} is widely adopted. 
Meanwhile,
self-instruct methods~\cite{wang-etal-2023-self-instruct,xu2023wizardlm,ouyang2022training} have been proposed to generate diverse and abundant instruction data, based on a small collection of seed instructions. In our work, we follow previous instruction tuning strategies in both tuning stages. For symbolic tasks, we construct instructions, 
covering the task and symbolic descriptions. For general tasks, we sample the instruction-tuning datasets (e.g., Flan~\cite{longpre2023flan}).

\paragraph{Symbol-centric Scenarios}
LLMs have dominated plenty of NL-centric tasks~\cite{rajpurkar2016squad,talmor-etal-2019-commonsenseqa,nallapati2016abstractive}, 
where NL is leveraged as the core interface for interaction, planning, and reasoning. 
But world knowledge is not purely represented by NL.
In fact, symbolic language is also of great significance in expressing abstract world knowledge~\cite{edwards2022translation,DBLP:conf/aaai/BevilacquaBN21,li-srikumar-2019-augmenting} and leveraging external tools~\cite{gao2023pal,DBLP:journals/corr/abs-2304-11477,pan2023logic}. 
Some concurrent works~\cite{xu2023lemur,yang2023harnessing} shift focus to the specific forms of symbols (e.g., code), either through prompting off-the-shelf LLMs or tuning on open-source LLMs. These efforts fail to lay a solid symbolic foundation, which is expected to grasp the interrelations among various symbolic forms. In our work, we explore the possibility of treating symbols in a unified manner and lay foundations to build balanced symbol and NL interfaces.

\section{Conclusion}
\vspace{-0.1cm}
This work pioneeringly proposes to enhance the LLM capability in symbol-centric tasks while preserving the performances on general tasks,
leading to balanced symbol and NL interfaces. 
To address the challenges of capturing symbol interrelations and maintaining a balance in general abilities, we tackle the problem from both data and framework perspectives.
Data-wise,
we include a collection of 34 text-to-symbol tasks to systematically explore underlying symbol relations.
Framework-wise,
we implement SFT in a novel two-stage manner to mitigate catastrophic forgetting.
Extensive experiments across three task settings (i.e., symbolic tasks, general tasks, and symbol+delegation tasks) demonstrate Symbol-LLM's superiority in harmonizing symbol- and NL-centric capabilities.
Moreover,
all models and resources will be made public to facilitate a broader range of research.

\section*{Limitations}
The insight of Symbol-LLM is to build a balanced symbol- and NL-centric interface for interaction and reasoning. We achieve it from both data (comprehensive symbolic collection to open-source) and framework (two-stage tuning to reduce forgetting) perspectives. It is expected to expand the scope of cutting-edge open-source LLMs largely and lay a new foundation for future work. Though plenty of experiments covering three settings are conducted, there still exist the following two directions for exploration: (1) The model's ability to self-correct or interact with environmental feedback in symbolic scenarios. It is also key to building language agents from language models. (2) Model size scaling to 70B or larger. As widely recognized, 7B or 13B LLMs are still not sufficient to build excellent language agents, especially when complex interaction is involved. Thus, it needs further exploration for the size scaling to the larger ones.

\bibliography{custom}
\bibliographystyle{acl_natbib}

\clearpage
\newpage
\appendix

\section{Key Contributions and Values}
\label{key_contributions}
Beyond the proposed data collection strategy and the two-stage tuning strategy, we would also like to emphasize the distinctive value of our work:

\noindent \textbf{(1) Insights on synergies between symbolic and natural language for LLMs.} Our motivation is sourced from the lack of symbolic foundation in off-the-shelf LLMs, which heavily focus on the NL capabilities. Equipping LLMs with a symbolic foundation brings (i) diverse expression of knowledge; (ii) great power in controlling robots and leveraging tools. Also, the synergies between symbolic and natural language can offer potential insights into transforming language models into language agents, which cover extensive code interaction and tool-using scenarios.

\noindent \textbf{(2) Thorough evaluations.} Quite different from previous works which merely focus on specific symbols and specific tasks, our work builds comprehensive evaluation studies under nearly 200 tasks. The evaluations are mainly divided into three folds: (i) symbolic tasks. (ii) general NL tasks. (iii) symbol+delegation tasks.

\noindent \textbf{(3) Open-sourcing effort for the LLM community.} Large numbers of recent works are based on closed-source LLMs (e.g., GPT-4). However, such methods are high-cost and inefficient, limiting custom training. Therefore, our efforts to open-source Symbol-LLM series as well as the comprehensive data collection are valuable to the community.

\section{Insights behind Symbol-LLM}
\label{research_insights}
This part provides key insights on (1) two-stage tuning strategy; (2) data scaling; and (3) intrinsic superiority.

\paragraph{Insight 1: Two-stage strategy.}

Despite their strong natural language processing abilities, open-source LLMs lack symbol-centric capabilities.
Naturally, we think about \textbf{how LMs grasp the foundation in addressing NL?} The rough pipeline of training LMs can be in two stages. The first stage focuses on pretraining with large-scale corpus and language modeling objectives. This stage disregards high-level skills (e.g., instruction-following and complex reasoning) and solely focuses on establishing the natural language foundation. In the second stage, the potential for instruction-following, and aligning human preference is enhanced through finetuning (or preference optimization).

Motivated by it, we adopt the two-stage training strategy. Initially, we lay a symbolic foundation, known as the Injection stage, where we solely inject symbolic knowledge into LLMs. Then, in the Infusion stage, we balance the NL-centric abilities and excite the symbolic ability through further finetuning on the mixture of datasets.

In fact, several concurrent works share similar paradigms with ours. For example, CodeLLaMA-Instruct is based on LLaMA-2-Chat, utilizing large-scale code data for pretraining and NL data for optimizing NL capability. CodeGeeX2 and Lemur are also similar works, primarily effective in code-related scenarios. Symbol-LLM extends the scope to symbolic language (as we stated, code is one of the specific symbols).

\paragraph{Insight 2: Data scaling.}

One of the interesting insights behind Symbol-LLM is to explore \textbf{how much data is sufficient to establish a symbolic foundation?}
Based on the previous works~\cite{zheng-etal-2022-robust}, mastering new tasks in large models primarily entails modifications to high-layer parameters. Therefore, we employ the Kolmogorov-Smirnov Test to reflect changes in LLM parameters, with the training data scaling.

\begin{figure}[t]
\large
\centering
\includegraphics[scale=0.58]{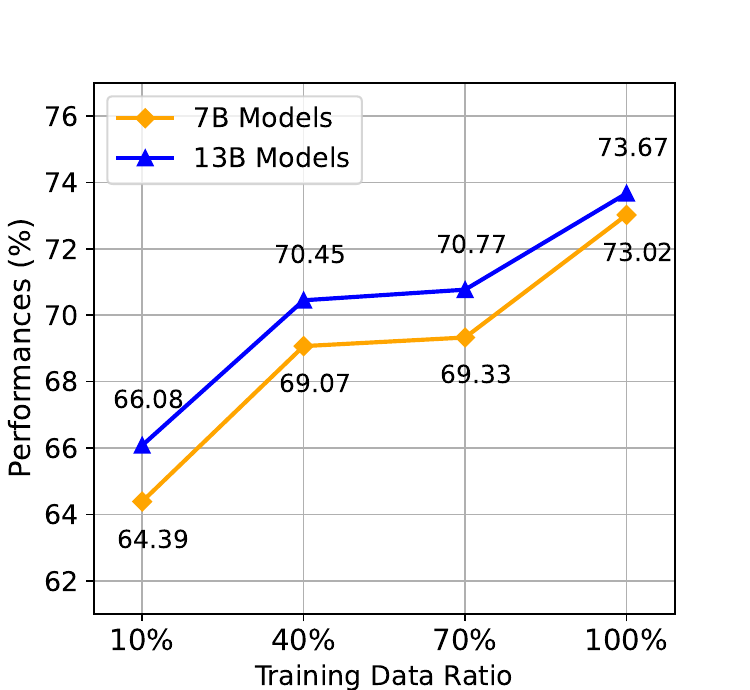}
\caption{Training data scaling.}
\label{scale}
\end{figure}

\begin{figure}[t]
\large
\includegraphics[scale=0.6]{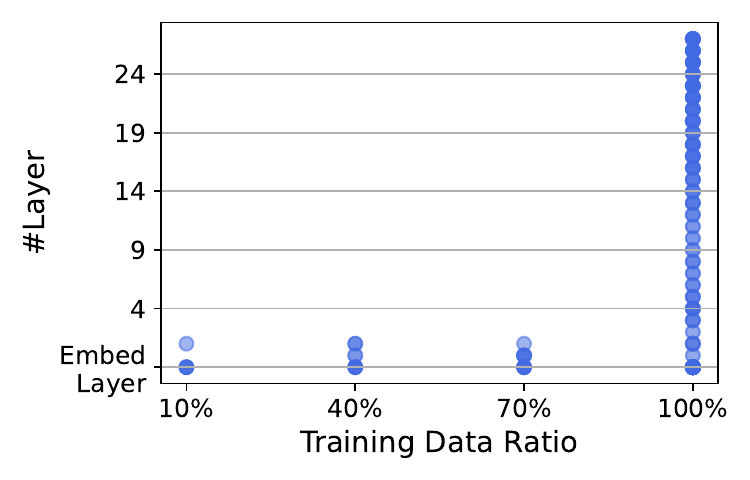}
\caption{KS-Test with data scaling. The scatter denotes the significant parameter changes (p-value<0.05). The darker color means a more significant change.}
\label{ks_scale}
\end{figure}

In Figure~\ref{scale}, we present the average performances on 34 symbolic tasks with data scaling. Specifically, we sample $\mathcal{D}_s$ in the \emph{Injection} stage at a ratio of 10\%, 40\%, 70\%, and 100\%. As observed, when symbolic training data surpasses 70\% (>620K), model performance breaks through the bottleneck of saturation.

Accordingly, the parameter changes layerwise are presented in Figure~\ref{ks_scale}. When training data scales from 70\% to 100\%, significant parameter changes emerge in higher layers. It is expected that scaling the symbolic training data can lay a new foundation of symbolic knowledge.

\paragraph{Insight 3: Intrinsic superiority.}

It is insightful to reveal the intrinsic superiority of Symbol-LLM over symbolic tasks. That is to say, \textbf{What is behind the superior performances of Symbol-LLM ?} We study it from \emph{Alignment and Uniformity} views. Details are discussed in Section~\ref{align_and_uniform} and Appendix ~\ref{appendix:align_uniform}.

\section{Details of Data Collection}
\label{appendix:data_collection}
In this section, detailed information on the data collection is attached, including both text-to-symbol task collection, and general task collection. 

\subsection{Text-to-symbol Task Collection}
\label{appendix:symbol_collection}
We provide a detailed illustration of the symbolic task collection, which consists of 34 different text-to-symbol generation tasks. 
They are categorized into 12 domains in Table~\ref{taskdetail}.

\begin{table*}[t]
\footnotesize
\centering
\resizebox{\linewidth}{!}{
\begin{tabular}{cl|cccccc}
    \toprule
    \textbf{Domains} & \textbf{Tasks} &\textbf{\# Train} &\textbf{\# Test} &\textbf{Sampled?} &\textbf{Access} &\textbf{Few-shot?} &\textbf{Original Source}   \\
    \hline
    \multirow{4}{*}{Planning} & Blocksworld &\multirow{4}{*}{37,600} &20 & &GPT-4+Evol &\checkmark &~\citet{DBLP:journals/corr/abs-2304-11477} \\
    & Termes & &20 & &GPT-4+Evol &\checkmark &~\citet{DBLP:journals/corr/abs-2304-11477} \\
    & Floortile & &20 & &GPT-4+Evol &\checkmark &~\citet{DBLP:journals/corr/abs-2304-11477}\\
    & Grippers & &20 & &GPT-4+Evol &\checkmark &~\citet{DBLP:journals/corr/abs-2304-11477} \\
    \hline
    \multirow{3}{*}{SQL} & Spider &\multirow{3}{*}{109,582} &1,034 & &Direct & &~\citet{DBLP:conf/emnlp/YuZYYWLMLYRZR18} \\
    & Sparc & &1,625 & &Direct & &~\citet{DBLP:conf/acl/YuZYTLLELPCJDPS19} \\
    & Cosql & &1,300 & &Direct & &~\citet{DBLP:conf/emnlp/YuZELXPLTSLJYSC19} \\
    \hline
    \multirow{3}{*}{KG / DB} & WebQSP &3,241 &1,639 & &Direct &\checkmark &~\citet{yih-etal-2016-value} \\
    & GrailQA &53,222 &6,463 & &Direct &\checkmark &~\citet{rogers2023qa} \\
    & CompWebQ &37,444 &3,531 & &Direct &\checkmark &~\citet{talmor-berant-2018-web} \\
    \hline
    \multirow{3}{*}{AMR} &AMR3.0 &68,778 &1,898 & &Direct &\checkmark &~\citet{knight2020amr} \\
    &AMR2.0 &45,436 &1,371 & &Direct &\checkmark &~\citet{knight2017amr} \\
    &BioAMR &7,150 &500 & &Direct &\checkmark &~\citet{banarescu2013abstract} \\
    \hline
    \multirow{2}{*}{Ontology} &Tekgen &11,219 &4,062 & &Direct &\checkmark &~\citet{agarwal2021knowledge} \\
    &Webnlg &3,415 &2,014 & &Direct &\checkmark &~\citet{gardent-etal-2017-creating} \\
    \hline
    \multirow{3}{*}{API} & MTOP &18,784 &500 &\checkmark &Direct &\checkmark &~\citet{li-etal-2021-mtop} \\
    & TOPv2 &149,696 &500 &\checkmark &Direct &\checkmark &~\citet{chen-etal-2020-low-resource} \\
    & NLmaps &21,657 &10,594 & &Direct &\checkmark &~\citet{lawrence-riezler-2018-improving} \\
    \hline
    \multirow{1}{*}{Command} & SCAN &25,990 &4,182 & &Direct+Evol &\checkmark &~\citet{lake2018generalization} \\
    \hline
    \multirow{5}{*}{Code} & NL2BASH &11,971 &746 & &Direct &\checkmark &~\citet{LinWZE2018:NL2Bash} \\
    &NL2RX &10,808 &1,000 &\checkmark &Direct &\checkmark &~\citet{locascio2016neural} \\
    &NL2Python &12,005 &500 &\checkmark &Direct &\checkmark &~\citet{husain2019codesearchnet}\\
    &NL2Java &11,978 &500 &\checkmark &Direct &\checkmark &~\citet{husain2019codesearchnet}\\
    &NL2Go &12,001 &500 &\checkmark &Direct &\checkmark &~\citet{husain2019codesearchnet}\\
    
    \hline
    \multirow{3}{*}{FOL} & FOLIO &2,006 &500 &\checkmark &Direct &\checkmark &~\citet{han2022folio}\\
    & MALLS &39,626 &1,000 &\checkmark &Direct &\checkmark &~\citet{yang2023harnessing} \\
    & LogicNLI &11,559 &2,373 &\checkmark &Direct &\checkmark &~\citet{tian-etal-2021-diagnosing} \\
    \hline
    \multirow{3}{*}{Visual} &GQA &36,086 &2,000 &\checkmark &Direct &\checkmark &~\citet{hudson2019gqa} \\
    & CLEVR &47,081 &2,000 &\checkmark &Direct+Evol &\checkmark &~\citet{johnson2017clevr}\\
    & Geometry3k &2,864 &601 & &Direct &\checkmark &~\citet{lu2021inter} \\
    \hline
    \multirow{3}{*}{Math} & GSM8K-Code &8,453 &100 &\checkmark &GPT-4 &\checkmark &~\citet{DBLP:journals/corr/abs-2110-14168} \\
    & AQUA-Code &31,144 &100 &\checkmark &GPT-4 &\checkmark &~\citet{DBLP:conf/acl/LingYDB17} \\
    & MATH-Code &4,426 &100 &\checkmark &GPT-4 &\checkmark &~\citet{DBLP:conf/nips/HendrycksBKABTS21} \\
    \hline
    \multirow{1}{*}{AI4Science} &CheBi-20 &35,629 &3,300 & &Direct &\checkmark &~\citet{edwards2022translation} \\
    \bottomrule
\end{tabular}}
\caption{Detailed illustrations of 34 text-to-symbol generation tasks. \emph{\# Train} and \emph{\# Test} represent the number of training and test samples respectively. \emph{Sampled?} means whether the test split is sampled from the original dataset. \emph{Access} is related to how we obtain the data, including directly from off-the-shelf benchmarks (Direct), prompting GPT-4 (GPT-4), and applying the symbol-evol strategy (Evol). \emph{Few-shot?} denotes whether few-shot samples are included. \emph{Original Source} is the citation of the original paper.}
\label{taskdetail}
\end{table*}

Note that the symbolic task collection includes but is not limited to the listed 34 tasks. To expand the diversity, we also consider some similar tasks. For example, we include some domain-specific NL-to-SQL tasks to provide diverse schema. The data is only used at the tuning stage but is not for a test. Thus, the whole collection (only count training samples) reaches $\sim$880K samples. All of them are leveraged in the first SFT stage.

Also, it is mentioned above that we sample parts of symbolic task collection in the second stage to reduce forgetting. For it, we uniformly sample each task domain with a ratio of 0.3, leading to a sampled collection of $\sim$260K.

\subsection{General Task Collection}
\label{appendix:general_collection}
In the second tuning stage, we include a collection of general instruction-tuning data to keep the LLM capability in some NL-centric settings and further improve the instruction-following capability of Symbol-LLM.

The general data collection contains $\sim570K$ samples, which are sourced from the following three parts:

(1) \textbf{Sampled Flan collection}~\cite{longpre2023flan} of 150K samples. We obtain the collection directly following Tulu~\cite{wang2023far}.

(2) \textbf{Code Alpaca collection}~\cite{codealpaca} of 20K samples. In fact, this collection is not in an NL-to-NL form as we expected. However, it stresses much on the instruction-following capabilities, which may help enhance the general ability of LLMs. Also, it is expected to act as the bridge between NL data and symbolic form (i.e., code in this case).

(3) \textbf{Sampled WizardLM collection}~\cite{xu2023wizardlm} of 143K samples. To further expand the diversity of our instruction-tuning collection, we leverage the evol-data from WizardLM.

\section{Data Format}
To support the instruction tuning, each piece of data $i$ in the training collection contains three parts, i.e., instruction $s_i$, input $x_i$, and output $y_i$. During the training process, instruction $s_i$ and input text $x_i$ are concatenated as the whole input sequence. The model is optimized to generate output $y_i$. One example in the \emph{FOLIO} dataset is as follows:

\noindent [Instruction] Transform the natural language sentence into first-order logic forms.

\noindent [Input] All people who regularly drink coffee are dependent on caffeine.

\noindent [Output] $\forall$x (Drinks(x) $\rightarrow$ Dependent(x))

In the implementation, we rewrite the instruction for each sample by prompting GPT-4, keeping the diversity of the instruction.

\section{Test Datasets and Benchmarks}
\label{appendix:test}
Our main experiments are conducted on both text-to-symbol tasks and general NL-centric tasks. Then this work also extends the scope to \emph{Symbol+Delegation} setting, which uses LLM to generate symbolic representation and delegate the reasoning process to the external solver. Such a setting satisfies our expectation to build a better symbol interface.

\subsection{Tests in Text-to-Symbol Generation Tasks}

\paragraph{Planning} These tasks involve controlling the robot to finish some tasks in the defined environments. The input is the natural language description of the initial states and the final goals, while the symbolic output in the Planning Domain Definition Language (PDDL) form can be executed by the symbolic planner. For the four settings, \emph{Blocksworld} involves stacking blocks in order. \emph{Termes} involves moving blocks to a specific position in the grid. \emph{Floortile} is to color the floors with the instructions. \emph{Grippers} is to gripper and move balls from room to room. We use the BLEU metric to measure the correctness of generated forms.

\paragraph{SQL} They cover three representative Text-to-SQL datasets, \emph{Spider}, \emph{Sparc} and \emph{Cosql}. Given the schema and the natural language query, the output is the corresponding SQL. We use the exact match as the metric.

\paragraph{KG / DB} It is similar to the Text-to-SQL tasks, which require generating the symbolic form of the query given the natural language question and schema. But \emph{WebQSP} and \emph{GrailQA} leverage the s-Expression form while \emph{CompWebQ} uses SPARQL format. We use the F1 metric for \emph{WebQSP} and the exact match metric for \emph{GrailQA} and \emph{CompWebQ}, following previous work~\cite{xie2022unifiedskg}.  

\paragraph{AMR} They are classical semantic parsing tasks, where the input sentence is parsed into an abstract syntax graph. We use the Smatch metric to measure the generated form on \emph{AMR3.0}, \emph{AMR2.0}, and \emph{BioAMR} datasets.

\paragraph{Ontology} It focuses on the domain of knowledge graph construction. Given the ontology (i.e, predefined relations or entities) and natural language sentence, it is required to output the triples. We employ F1 scores introduced in~\cite{mihindukulasooriya2023text2kgbench} to measure the performances on \emph{Tekgen} and \emph{WebNLG}.

\paragraph{API} These tasks require the output of the API calling form based on the natural language query. \emph{MTOP} and \emph{TOPv2} cover various domains like controlling the music player, and setting alarms. \emph{NLMAPS} focuses on calling the maps.

\paragraph{Command} \emph{SCAN} involves outputting action sequences based on the commands to control robots. The exact match metric is used to measure the generation accuracy.

\paragraph{Code} It involves five representative programming languages, including \emph{Bash}, \emph{Regular Expression}, \emph{Python}, \emph{Java} and \emph{GO}. They are tested with the BLEU metric.

\paragraph{FOL} It covers three datasets in NL-to-FOL domain, that is \emph{FOLIO}, \emph{MALLS} and \emph{LogicNLI}. Logic Equivalence (LE) is leveraged as the metric, following~\cite{yang2023harnessing}.

\paragraph{Visual} Three multi-modal question answering datasets \emph{GQA}, \emph{Clevr} and \emph{Geometry3K} are included for test. In these scenarios, we only focus on the natural language parts and transform the natural language query into function symbol forms. The exact match metric is used to measure the performances.

\paragraph{Math} As we discussed, transforming the natural language question into Python code is one of the faithful ways to solve math problems. Hence, we measure the accuracy of the generated Python code with the BLEU metric. The ground-truth code is derived by prompting GPT-4, where the ones that can execute the correct answer are preserved.

\paragraph{AI4Science} In \emph{CheBi} dataset, the model is required to generate the correct molecular formula given the natural language descriptions. Exact match metric is used for measure.


\subsection{Tests in General Tasks}

\paragraph{MMLU} It covers 57 tasks including different subjects STEM, humanities, social sciences, and others. Our evaluations are based on~\cite{hendryckstest2021}.

\paragraph{Big Bench Hard} The benchmark is designed for testing LLM capability in challenging reasoning tasks. We select 21 tasks in BBH for the test, based on Open-LLM-Leaderboard\footnote{\url{https://huggingface.co/spaces/HuggingFaceH4/open_llm_leaderboard}}. 


\subsection{Tests in Symbol+Delegation Setting}
\label{appendix:symbol+delegation}
\paragraph{Math Reasoning} We generate Python code with Symbol-LLM and use Python interpreter as the delegation. The datasets include GSM8K~\cite{DBLP:journals/corr/abs-2110-14168}, MATH~\cite{DBLP:conf/nips/HendrycksBKABTS21}, GSM-Hard~\cite{gao2023pal}, SVAMP~\cite{patel2021nlp}, Asdiv~\cite{miao-etal-2020-diverse}, AddSub~\cite{hosseini-etal-2014-learning}, SingleEQ~\cite{roy2015reasoning}, SingleOP~\cite{roy2015reasoning} and MultiArith~\cite{roy2015solving}. The former two datasets are in-domain, while the latter seven datasets are under OOD settings. 

Note that MATH dataset includes various ground-truth answer formats (e.g., with diverse units), thus it is difficult to parse the correct values to evaluate the LLMs. Hence, we use manually-crafted templates to derive the ground-truth values, leading to around 4,000 samples for test.

\paragraph{Symbolic Reasoning} Same as math reasoning, we use Python code + Python interpreter to solve the problems. Two OOD tasks are used for test, i.e., Colored Objects~\cite{DBLP:conf/acl/SuzgunSSGTCCLCZ23} and Last Letter Concatenation~\footnote{Test data is based on: \url{https://huggingface.co/datasets/ChilleD/LastLetterConcat}}.

\paragraph{Logical Reasoning} We take three representative datasets into consideration, i.e., FOLIO~\cite{han2022folio}, ProofWriter~\cite{tafjord2021proofwriter} and ProntoQA~\cite{saparov2022language}. We follow the strategy proposed in ~\cite{pan2023logic} to conduct the reasoning. Detailedly, for FOLIO, we generate FOL representations first and delegate the solution to the FOL solver. For ProofWriter and ProntoQA tasks, we generate logic programming language and delegate the reasoning to \emph{Pyke} expert system. 

\paragraph{Robotic Planning} For robotic planning tasks, we transform the natural language description into PDDL and use fastdownward~\cite{helmert2006fast} as the symbolic solver. Besides the four datasets mentioned in text-to-symbol generation tasks, we also employ two OOD datasets into account, i.e. Barman and Tyreworld.

\paragraph{Visual Question Answering} We further extend the application scope of Symbol-LLM to the multi-modal domain and test on Geometry3K dataset~\cite{lu2021inter} for illustration. But we only concentrate on the processing of the NL part. Detailed, we parse the natural language sentence into logic forms and rely on the baseline method~\cite{lu2021inter} to conduct the multi-modal reasoning.

\section{Experimental Settings}
\label{appendix:exp_settings}


In the implementation, this work fully finetunes the models with 8 * A100 (80GB) with maximum sequence length set to 4,096. The tuning process is optimized and accelerated by deepspeed zero3 and FlashAttention2. We leverage the AdamW optimizer with a learning rate of 2e-5 for both \emph{Injection} and \emph{Infusion} stages. The learning rate schedular is set to \emph{Linear}. The epoch number is set to 1 for both stages. In the \emph{Injection} stage, the model weights are initialized from LLaMA-2-Chat and the tuned model is named \slmb. In the \emph{Infusion} stage, we initialize the model from \slmb and obtain \slmi at last. These settings are consistent for both 7B and 13B variants.

The inference process is conducted under a single GPU of A100 (80GB) with greedy search (beam size=1).

For a comprehensive evaluation, we include the following strong baselines. They are categorized into \emph{Close-source} and \emph{Open-source} ones:

\paragraph{Close-source Baselines}

\begin{itemize}
    \item \textbf{GPT-3.5} We access OpenAI API to call the model. Specifically, \texttt{GPT-3.5-turbo} version is employed for evaluation across a wide range of tasks.
    
    \item \textbf{Claude-1} We access Anthropic API to call the model. We select \texttt{Claude-instant-1.2} version for evaluation.
\end{itemize}

\paragraph{Open-source Baselines}

\begin{itemize}
    \item \textbf{LLaMA-2-Chat} Since \slm is initialized from LLaMA-2-Chat, we include it as the baseline. In general, LLaMA-2-Chat series is regarded as an excellent NL-centric interface for interaction and reasoning, which exhibits great performance on vast NL tasks.

    \item \textbf{Single SFT} We conduct SFT on LLaMA-2-Chat models for tasks in one specific domain. The obtained models can fully overfit the single domain, thus serving as a strong baseline for comparison.

    \item \textbf{CodeLLaMA-Instruct} Based on the origin LLaMA-2 models, the CodeLLaMA series is continually pretrained and finetuned with code data. Considering code is one of the specific symbols in our work, we include it as one of the strong baselines. For balanced capabilities in general tasks, we leverage \emph{CodeLLaMA-Instruct} for evaluations.
\end{itemize}

\begin{figure*}[t]
    \centering
    \begin{minipage}[t]{0.22\linewidth}
        \large
        \raggedleft
        \includegraphics[scale=0.29]{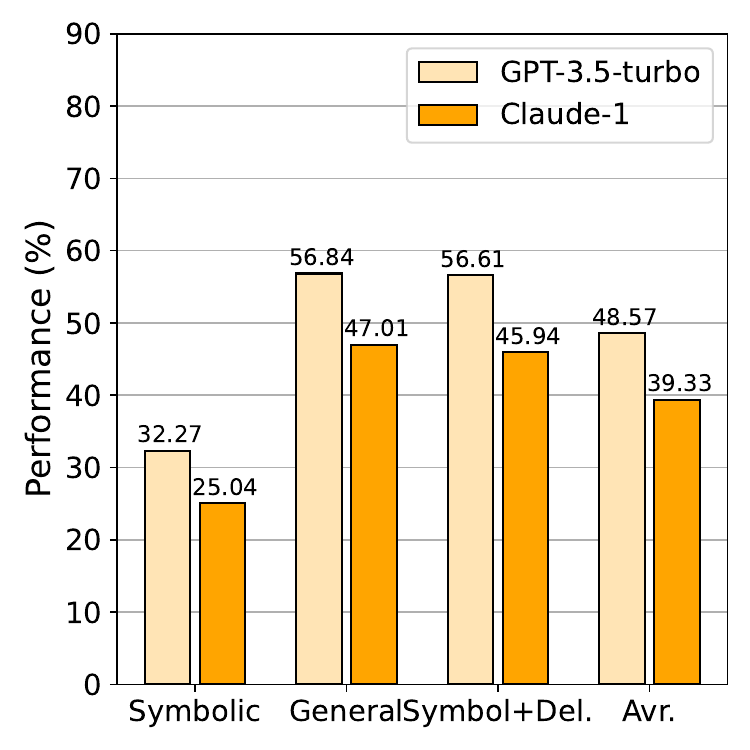}
        \subcaption{Close-source}
        \label{overall_close}
    \end{minipage}
    \begin{minipage}[t]{0.39\linewidth}
        \large
        \centering
        \includegraphics[scale=0.29]{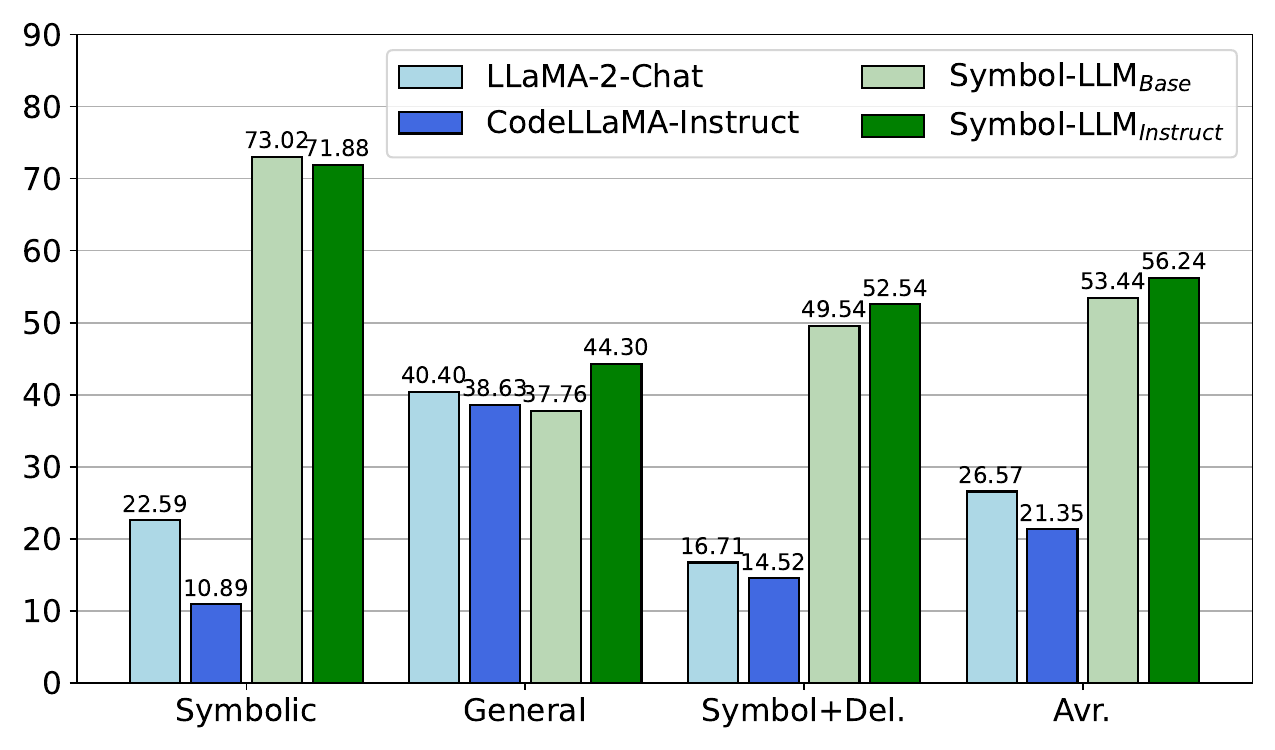}
        \subcaption{Open-source 7B}
        \label{overall_7b}
    \end{minipage}
    \begin{minipage}[t]{0.37\linewidth}
        \large
        \centering
        \includegraphics[scale=0.29]{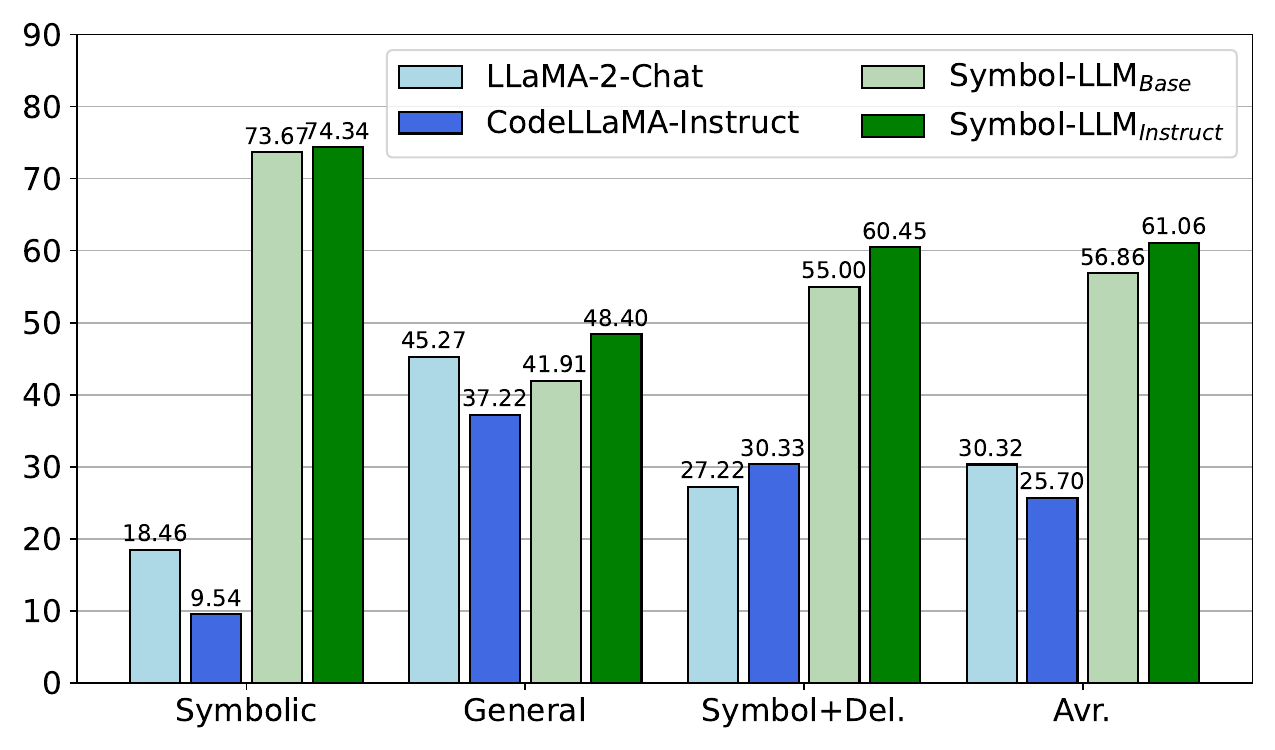}
        \subcaption{Open-source 13B}
        \label{overall_13b}
    \end{minipage}
    \caption{Overall results comparison. We report the performances on close-source, open-source 7B as well as open-source 13B LLMs. The results on symbolic tasks, general tasks, symbol+delegation tasks and the average ones are included.}
    \label{overall}
\end{figure*}

\section{Overall Performances}
\label{appendix:overall_performances}
Figure~\ref{overall} presents the overall performance comparison among baseline models.
It intuitively demonstrates the obvious advantages of \slm on the wide range of tasks. Also, it supports our claim to make \slm a balanced foundation LLM on symbols and NL.

\section{Results on OOD Symbolic Tasks}
\label{symbol_ood}
We supplement the experiments and compare the performances on 12 new OOD symbolic tasks. These OOD tasks can be divided into two folds: (i) novel tasks with seen symbolic forms (i.e., out-of-distribution); (ii) novel symbolic forms (i.e., out-of-domain). The results are presented in Table~\ref{exp_symbolic_ood}.

\begin{table*}[t]
\centering
\small
\begin{tabular}{l|cccc}
    \toprule
    \multirow{2}{*}{\textbf{Tasks}} &LLaMA-2-Chat &Symbol-LLM &LLaMA-2-Chat &Symbol-LLM \\
    &\makecell[c]{7B} &\makecell[c]{7B} &\makecell[c]{13B} &\makecell[c]{13B}\\
    \hline
    \multicolumn{5}{c}{\cellcolor{gray!20} Out-of-distribution Tasks} \\
    \hline
    Tyreworld &23.12 &\textbf{33.30} &23.85 &\textbf{79.18} \\
    WikiSQL &21.05 &\textbf{73.75} &34.86 &\textbf{69.83} \\
    WikiTQ &3.50 &\textbf{16.97} &7.50 &\textbf{15.31} \\
    SVAMP &22.00 &\textbf{72.80} &45.00 &\textbf{76.80} \\
    Asdiv &25.86 &\textbf{75.76} &46.61 &\textbf{79.01} \\
    \hline
    \multicolumn{5}{c}{\cellcolor{gray!20} Out-of-domain Tasks} \\
    \hline
    NL2JS &16.36 &\textbf{16.64} &14.35 &\textbf{17.60} \\
    NL2PHP &9.27 &\textbf{23.38} &22.84 &\textbf{26.31} \\
    H-Termes &39.39 &\textbf{63.87} &39.64 &\textbf{69.08} \\
    H-Floortile &47.66 &\textbf{76.88} &63.57 &\textbf{79.57} \\
    H-Grippers &79.52 &\textbf{98.96} &55.87 &\textbf{99.00} \\
    H-Scan &0.00 &\textbf{91.25} &0.00 &\textbf{95.34} \\
    H-Clevr &0.00 &\textbf{98.52} &0.00 &\textbf{98.02} \\
    \bottomrule
\end{tabular}
\caption{Results on OOD symbolic tasks. Tasks starting with `H-' mean the symbolic forms are automatically modified to construct the much harder scenarios and imitate the user-defined settings.}
\label{exp_symbolic_ood}
\end{table*}

From the results, Symbol-LLM shows consistent and significant superiority, compared with LLaMA-2-Chat.

\section{Results on Extensive General Tasks}
\label{appendix:exp_general_tasks}

\begin{table}[t]
\centering
\footnotesize
\begin{tabular}{l|cc}
    \toprule
    \textbf{Tasks} &LLaMA-2-Chat$\dagger$ &Symbol-LLM \\
    \hline
    \multicolumn{3}{c}{\cellcolor{gray!20} Examinations} \\
    \hline
    AGI-Eval &\textbf{28.50} &27.55 \\
    C-Eval &31.90 &\textbf{34.96} \\
    GaokaoBench &\textbf{16.10} &13.37 \\
    ARC-c &54.90 &\textbf{61.69} \\
    \hline
    \multicolumn{3}{c}{\cellcolor{gray!20} Knowledge} \\
    \hline
    BoolQ &\textbf{81.30} &77.00 \\
    CommonsenseQA &\textbf{69.90} &59.21 \\
    TrivialQA &\textbf{46.40} &40.90 \\
    NaturalQuestions &\textbf{19.60} &16.48 \\
    \hline
    \multicolumn{3}{c}{\cellcolor{gray!20} Understanding} \\
    \hline
    OpenbookQA &74.40 &\textbf{79.20} \\
    XSUM &20.80 &\textbf{33.29} \\
    LAMBADA &66.90 &\textbf{70.10} \\
    C3 &49.80 &\textbf{52.82} \\
    \hline
    \multicolumn{3}{c}{\cellcolor{gray!20} Reasoning} \\
    \hline
    CMNLI &36.10 &\textbf{55.43} \\
    OCNLI &36.40 &\textbf{48.10} \\
    Ax-b &58.50 &\textbf{62.68} \\
    Ax-g &51.70 &\textbf{64.61} \\
    Hellaswag &\textbf{74.10} &53.10 \\
    SIQA &55.40 &\textbf{69.60} \\
    MBPP &17.60 &\textbf{22.80} \\
    ReCoRD &22.50 &\textbf{39.49} \\
    \bottomrule
\end{tabular}
\caption{Results on extensive general tasks. The evaluations are based on OpenCompass~\cite{2023opencompass}. $\dagger$ denotes the model results directly derived from the leaderboard. }
\label{general_app}
\end{table}

In Table~\ref{general_app}, we select extensive general tasks for comparison. According to OpenCompass~\cite{2023opencompass}, these tasks are divided into several categories, covering \emph{Examinations}, \emph{Knowledge}, \emph{Understanding} and \emph{Reasoning}. Considering the computation cost, we only report the performances of 7B models. The major takeaways are as follows:

\paragraph{Symbol-LLM demonstrates better overall performances compared with LLaMA-2-Chat.} Generally speaking, Symbol-LLM wins more of the tasks than LLaMA-2-Chat. Such superiority is consistent with the findings in MMLU and BBH benchmarks in Table~\ref{general}. It illustrates that Symbol-LLM can serve as a solid foundational model, significantly enhancing its symbolic capabilities while maintaining its generality.

\paragraph{Optimization empowers Symbol-LLM with improved understanding and reasoning abilities.} Among the four task categories, Symbol-LLM is particularly better at understanding and reasoning, beating LLaMA-2-Chat on almost all tasks. Such findings are intuitive because text-to-symbol can be regarded as an abstract form of NL, which enriches the understanding abilities of the model. Meanwhile, the generation of some symbolic forms (e.g., code) involves the implicit reasoning process, which is actually similar to the chain-of-thought strategy. To this end, the superior reasoning capability is within our expectations.

\section{Results on Symbol+Delegation Setting}
\label{appendix:exp_symbol_delegation}

In the main paper, we present the results of math reasoning under the \emph{Symbol+Delegation} paradigm. Next, we will provide the remaining 5 scenarios, i.e., symbolic reasoning~(\ref{appendix_symbolic_reasoning}), logical reasoning~(\ref{appendix:logical_reasoning}), robotic planning~(\ref{appendix:robotic_planning}), visual question answering~(\ref{appendix:visual_question_answering}) and table question answering~(\ref{appendix:table_question_answering}).

\subsection{Symbolic Reasoning}
\label{appendix_symbolic_reasoning}

\begin{table}[t]
\centering
\footnotesize
\resizebox{\linewidth}{!}{
\begin{tabular}{lc|cc}
    \toprule
    \textbf{Models} & \textbf{Del.} &\textbf{ColoredObject} &\textbf{LastLetter} \\
    \hline
    \multicolumn{2}{l|}{\textbf{Is OOD Setting}} &\color{ao(english)}\checkmark &\color{ao(english)}\checkmark\\
    \hline
    \multicolumn{4}{c}{\cellcolor{gray!20} Close-source LLMs} \\
    \hline
    GPT-3.5-turbo &\color{ao(english)}\checkmark &12.45 &94.00 \\
    Claude-1 &\color{ao(english)}\checkmark &46.05 &90.67 \\
    \hline
    \multicolumn{4}{c}{\cellcolor{gray!20} Open-source LLMs (7B)}
    \\
    \hline
    LLaMA-2-Chat &\color{ao(english)}\checkmark &\textbf{28.70} &0.00 \\
    CodeLLaMA-Instruct &\color{ao(english)}\checkmark &4.60 &0.00 \\
    \textbf{\slmb} &\color{ao(english)}\checkmark &22.65 &90.67  \\
    \textbf{\slmi} &\color{ao(english)}\checkmark &\underline{25.50} &\textbf{96.67}  \\
    \hline
    \multicolumn{4}{c}{\cellcolor{gray!20} Open-source LLMs (13B)}
    \\
    \hline
    LLaMA-2-Chat &\color{ao(english)}\checkmark &30.35 &0.00 \\
    CodeLLaMA-Instruct &\color{ao(english)}\checkmark &1.35 &0.00 \\
    \textbf{\slmb} &\color{ao(english)}\checkmark &\textbf{36.35} &94.00 \\
    \textbf{\slmi} &\color{ao(english)}\checkmark &\underline{34.00} &\textbf{96.67}  \\
    \bottomrule
\end{tabular}}
\caption{Results on Symbolic Reasoning.
}
\label{symbolic}
\end{table}

In symbolic tasks, we also adopt the Python code as the generated symbolic representations, and leverage a Python interpreter to conduct the reasoning. 
Two representative tasks, \emph{Colored Objects} and \emph{Last Letter Concatenation} are selected for testing under the zero-shot setting.

From the results in Table~\ref{symbolic}, the Symbol-LLM series are competitive in both tasks. 
Notably,
even Symbol-LLM-7B shows over 10\% superiority over GPT-3.5-turbo in \emph{Colored Object} task. 
It is worth noticing that LLaMA-2-Chat models underperform consistently in \emph{Last Letter} task. 
Since samples in this dataset share similar forms, 
the model tends to fail if the model does not master the techniques required for solving it.

\subsection{Logical Reasoning}
\label{appendix:logical_reasoning}

\begin{table}[t]
\centering
\footnotesize
\resizebox{\linewidth}{!}{
\begin{tabular}{lc|ccc}
    \toprule
    \textbf{Models} & \textbf{Del.} &\textbf{FOLIO} & \textbf{ProofWriter} &\textbf{PrOntoQA} \\
    \hline
    \multicolumn{2}{l|}{\textbf{Is OOD Setting}} &\color{red}\usym{2717} &\color{red}\usym{2717} &\color{ao(english)}\checkmark\\
    \hline
    \multicolumn{5}{c}{\cellcolor{gray!20} Close-source LLMs} \\
    \hline
    GPT-3.5-turbo &\color{ao(english)}\checkmark &44.61 &29.00 &52.00 \\
    Claude-1 &\color{ao(english)}\checkmark &37.25 &35.83 &55.80 \\
    Logic-LM (SOTA) &\color{ao(english)}\checkmark &61.76 &70.11 &93.20 \\
    \hline
    \multicolumn{5}{c}{\cellcolor{gray!20} Open-source LLMs (7B)}
    \\
    \hline
    LLaMA-2-Chat &\color{ao(english)}\checkmark &34.80 &34.83 &50.00 \\
    CodeLLaMA-Instruct &\color{ao(english)}\checkmark &32.84 &32.50	&50.20 \\
    \textbf{\slmb}&\color{ao(english)}\checkmark &\underline{46.08} &\textbf{76.50} &\underline{55.60} \\
    \textbf{\slmi} &\color{ao(english)}\checkmark &\textbf{49.02} &\underline{76.33} &\textbf{57.20} \\
    \hline
    \multicolumn{5}{c}{\cellcolor{gray!20} Open-source LLMs (13B)}
    \\
    \hline
    LLaMA-2-Chat &\color{ao(english)}\checkmark &33.33 &35.83 &49.20 \\
    CodeLLaMA-Instruct &\color{ao(english)}\checkmark &32.84 &34.00	&50.00 \\
    \textbf{\slmb} &\color{ao(english)}\checkmark &\underline{33.82} &\textbf{76.33} &48.40 \\
    \textbf{\slmi} &\color{ao(english)}\checkmark &\textbf{35.29} &\underline{75.50} &\textbf{53.60} \\
    \bottomrule
\end{tabular}}
\caption{Results on Logical Reasoning.
All results are obtained under the one-shot setting.}
\label{logic}
\end{table}

In logical reasoning tasks, we take three tasks into consideration, i.e., FOLIO, ProofWriter and ProntoQA. For the FOLIO task, Symbol-LLM first transforms the natural language into FOL forms and delegates the solution to the FOL solver. ProofWriter and ProntoQA are represented in logic programming language and integrate \emph{Pyke} expert system for deductive reasoning.

Results are listed in Table \ref{logic}. Symbol-LLM-7B series performs relatively better than 13B counterparts. Among all three tasks, the Symbol-LLM-7B series outperforms GPT-3.5-turbo with large advantages. In comparison with the SOTA model Logic-LM, which is based on off-the-shelf LLMs, Symbol-LLM also wins the ProofWriter tasks, with 5\%-6\% improvements.

\subsection{Robotic Planning}
\label{appendix:robotic_planning}
\begin{table*}[t]
\centering
\footnotesize
\resizebox{0.9\textwidth}{!}{
\begin{tabular}{lc|cccccc}
    \toprule
    \textbf{Models} & \textbf{Del.} &\textbf{Blocksworld} & \textbf{Termes} &\textbf{Floortile} & \textbf{Grippers}  & \textbf{Barman} & \textbf{Tyreworld}\\
    \hline
    \multicolumn{2}{l|}{\textbf{Is OOD Setting}} &\color{red}\usym{2717} &\color{red}\usym{2717} &\color{red}\usym{2717} &\color{red}\usym{2717} &\color{ao(english)}\checkmark &\color{ao(english)}\checkmark\\
    \hline
    \multicolumn{8}{c}{\cellcolor{gray!20} Close-source LLMs} \\
    \hline
    GPT-3.5-turbo &\color{ao(english)}\checkmark &55.00 &0.00	&0.00 &100.00 &95.00 &30.00 \\
    Claude-1 &\color{ao(english)}\checkmark &55.00 &0.00 &0.00 &85.00 &50.00 &5.00 \\
    \hline
    \multicolumn{8}{c}{\cellcolor{gray!20} Open-source LLMs (7B)}
    \\
    \hline
    LLaMA-2-Chat &\color{ao(english)}\checkmark &5.00 &0.00 &0.00 &5.00 &0.00 &0.00 \\
    LLaMA-2-Chat SFT &\color{ao(english)}\checkmark &75.00 &\textbf{100.00} &0.00 &0.00 &0.00 &0.00 \\
    CodeLLaMA-Instruct &\color{ao(english)}\checkmark &5.00	&0.00 &0.00	&20.00 &0.00 &0.00 \\
    \textbf{\slmb} &\color{ao(english)}\checkmark &\underline{90.00} &\textbf{100.00} &\underline{5.00} &\underline{15.00} &0.00 &0.00 \\
    \textbf{\slmi} &\color{ao(english)}\checkmark &\textbf{100.00} &50.00 &\textbf{20.00}	&\textbf{20.00} &0.00 &\textbf{5.00} \\
    \hline
    \multicolumn{8}{c}{\cellcolor{gray!20} Open-source LLMs (13B)}
    \\
    \hline
    LLaMA-2-Chat &\color{ao(english)}\checkmark &0.00 &0.00 &0.00 &\textbf{45.00} &\textbf{50.00} &5.00 \\
    LLaMA-2-Chat SFT &\color{ao(english)}\checkmark &70.00 &\textbf{100.00} &\textbf{25.00} &10.00 &0.00 &0.00 \\
    CodeLLaMA-Instruct &\color{ao(english)}\checkmark &5.00 &0.00 &0.00 &0.00 &0.00 &0.00 \\
    \textbf{\slmb} &\color{ao(english)}\checkmark &\underline{90.00} &\textbf{100.00} &0.00 &30.00 &0.00 &\underline{10.00} \\
    \textbf{\slmi} &\color{ao(english)}\checkmark &\textbf{100.00} &90.00 &\textbf{25.00} &\textbf{45.00} &\underline{20.00} &\textbf{35.00} \\
    \bottomrule
\end{tabular}}
\caption{Results on Robotic Planning. The evaluation is under the one-shot setting.
}
\label{robot}
\end{table*}

In the field of robotic planning,
\slm first transforms the natural language description into PDDL forms and relies on the fast downward solver to give the faithful action sequence.

In total, 
we select 6 different robotic settings to verify the proposed method. 
Results are presented in Table~\ref{robot}. 
Among four in-domain tasks, \slm performs pretty well compared with strong baselines, achieving the best results in most cases. Even with GPT-3.5-turbo and Claude-1, both our 7B and 13B series win 3 (out of 4) tasks. However, it struggles a lot in OOD tasks. Only in \emph{Tyreworld} scenario, 
\slmi-13B
achieves the best result, beating all close-source and open-source baselines. 
It is required to state that these selected robotic planning tasks are very challenging, given the length and rigor requirements of the generated programming language. Even close-source LLMs fail in some scenarios. Therefore, we argue it is still an open question for future studies.

\subsection{Visual Question Answering}
\label{appendix:visual_question_answering}
\begin{figure}[t]
\centering
\includegraphics[scale=0.43]{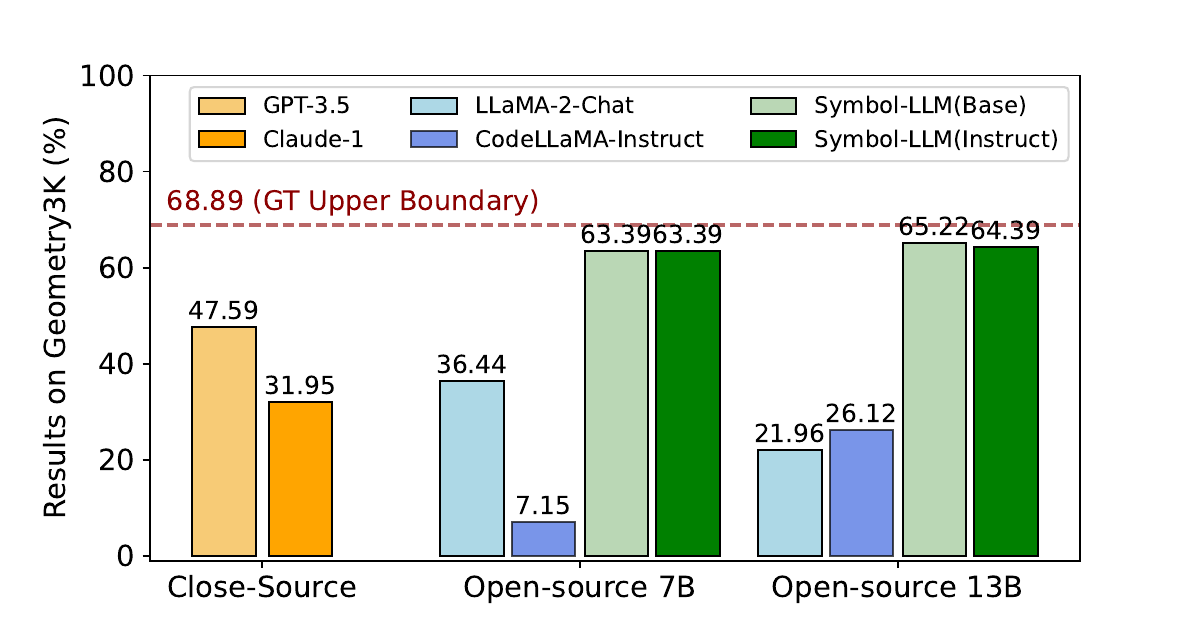}
\caption{Performances on Geometry3k task.}
\label{geometry}
\end{figure}

We also explore our potential in the multi-modal scenario. Geometry question answering is selected as the task for the test. Note that the understanding of image is not within our scope, we only focus on the text part and transform the natural language query into logical forms. Then the solution is delegated to the off-the-shelf baseline methods. Comparison results are shown in Figure~\ref{geometry}.

The top red bar means the performances using the annotated logic forms from the text. Note that since the utilized delegation method is a neural-based baseline just for a simple evaluation, the upper boundary does not represent the boundary of this task. From the figure, Symbol-LLM variants are approaching it and significantly outperform all the other baselines.

\subsection{Table Question Answering}
\label{appendix:table_question_answering}

\begin{table}[t]
\centering
\footnotesize
\resizebox{\linewidth}{!}{
\begin{tabular}{lc|cc}
    \toprule
    \textbf{Models} & \textbf{Del.} &\textbf{WikiSQL} & \textbf{WikiTQ} \\
    \hline
    \multicolumn{2}{l|}{\textbf{Is OOD Setting}} &\color{ao(english)}\checkmark &\color{ao(english)}\checkmark \\
    \hline
    \multicolumn{4}{c}{\cellcolor{gray!20} Close-source LLMs} \\
    \hline
    GPT-3.5-turbo &\color{ao(english)}\checkmark &28.49 &11.58 \\
    Claude-1 &\color{ao(english)}\checkmark &26.79	&8.79 \\
    \hline
    \multicolumn{4}{c}{\cellcolor{gray!20} Open-source LLMs (7B)}
    \\
    \hline
    LLaMA-2-Chat &\color{ao(english)}\checkmark &21.05 &3.50 \\
    CodeLLaMA-Instruct &\color{ao(english)}\checkmark &20.18 &2.88 \\
    \textbf{\slmb} &\color{ao(english)}\checkmark &\underline{70.88} &\textbf{17.15} \\
    \textbf{\slmi} &\color{ao(english)}\checkmark &\textbf{73.75} &\underline{16.97} \\
    \hline
    \multicolumn{4}{c}{\cellcolor{gray!20} Open-source LLMs (13B)}
    \\
    \hline
    LLaMA-2-Chat &\color{ao(english)}\checkmark &34.86 &7.50 \\
    CodeLLaMA-Instruct &\color{ao(english)}\checkmark &33.15 &6.70 \\
    \textbf{\slmb} &\color{ao(english)}\checkmark &\textbf{71.69} &\textbf{17.31} \\
    \textbf{\slmi} &\color{ao(english)}\checkmark &\underline{69.83} &\underline{15.31} \\
    \bottomrule
\end{tabular}}
\caption{Results on Table Question Answering.
}
\label{tableqa}
\end{table}

Table (or database) question answering is also a hot topic in recent years. 
Thus, we select two OOD tasks WikiSQL and WikiTQ for evaluations. 
The natural language query is first transformed into an SQL query and it is executed by an SQL solver over the given tables or databases under the zero-shot setting. We report experimental results in Table~\ref{tableqa}. 

Symbol-LLM series are consistently superior to all open-source and close-source baselines, with over 40\% margins in WikiSQL and 3\%$\sim$14\% advantages in WikiTQ.

\section{Supplementary Experiments}
\label{appendix:supplementary_exp}

\subsection{Tuning Design Choices}
\label{appendix:tuning_design_choice}
\begin{figure}[t]
\centering
\includegraphics[scale=0.68]{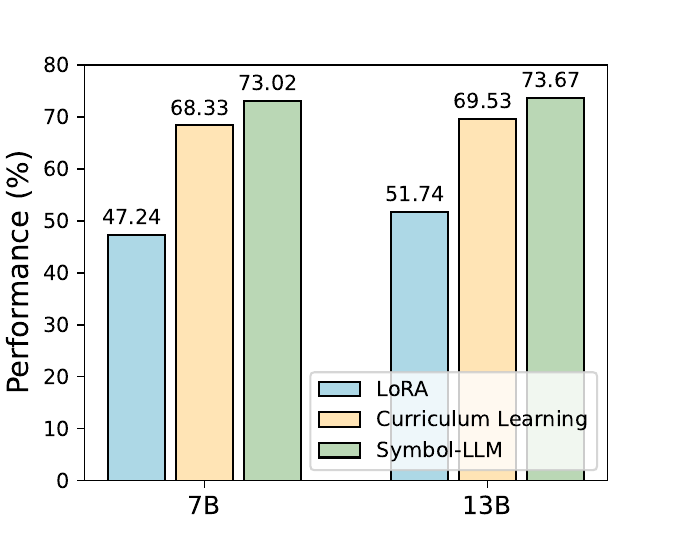}
\caption{Comparison between different tuning design choices.}
\label{tuning_design}
\end{figure}

Since our motivation is to build a symbolic foundation for current LLMs, we select the full fine-tuning strategy in the implementation. This part provides a comparison with different tuning design choices: (1) LoRA fine-tuning and (2) curriculum learning. In Figure~\ref{tuning_design}, the performances on 34 symbolic tasks are presented.

\paragraph{LoRA tuning} To inject new symbolic knowledge and lay a symbolic foundation for LLM, adapter tuning (e.g., LoRA) is not sufficient compared with the full fine-tuning design.

\paragraph{Curriculum learning} Based on the performances of LLaMA-2-Chat on these 34 tasks, we rank the difficulty level of tasks, forming the training order of curriculum learning (from easy to hard tasks). From the results, curriculum learning performs 4\%-5\% worse than our design choice.

\subsection{Comparison with Single Domain SFT}
\label{appendix:single_sft}

\begin{figure*}[t]
\large
\centering
\includegraphics[scale=0.6]{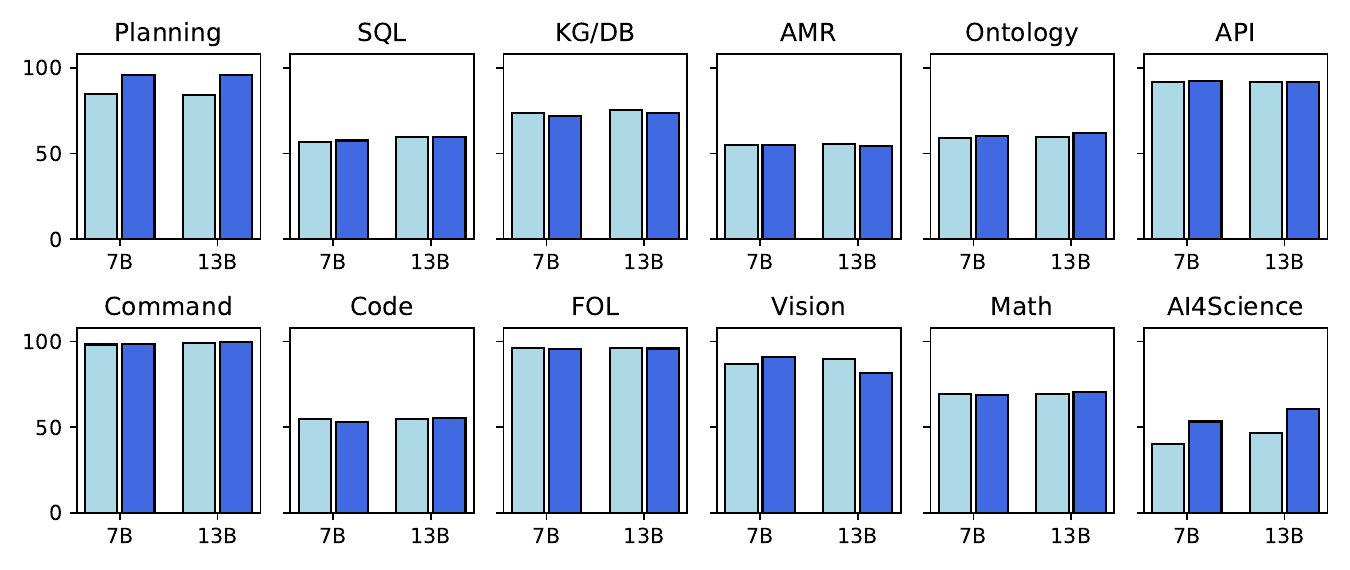}
\caption{Comparison between single SFT and unified SFT.}
\label{single}
\end{figure*}

As discussed above, 
one of our hypotheses is that various symbols share underlying interrelations, though they are in quite different forms. Thus, we expect that the learning of symbolic knowledge will mutually benefit each other if they are treated in a unified manner.

We present the comparison results between single SFT and unified SFT in Figure \ref{single}. The light blue bar denotes the single domain SFT while the dark blue one is the unified SFT.
We categorize the text-to-symbol generation tasks into 12 task domains, according to their similarity in symbolic forms. Within one domain, all tasks are tuned together and the performances on test splits are averaged as the single SFT results.
To reduce the effect of the tuning strategy, we utilize the \slmb model to measure the results on the unified SFT setting. 
Each sub-figure in Figure~\ref{single} corresponds to one specific domain.

In most domains, unified SFT is superior to single-domain SFT. Larger gains are observed in some uncommon symbolic forms, such as PDDL for planning tasks and molecular formulas in AI for Science scenarios. It presents the possibility that unified SFT on various symbols may help extend the model coverage to low-resource cases. It is also worth noting that in some cases, single-domain SFT performs a little better than \slmb. This is because purely overfitting on specific symbolic forms with powerful LLMs is usually easy to get promising results.

\subsection{Extrapolating to New Symbols}
\label{appendix:extrapolation}
In the above section, we introduce \emph{symbol-evol} strategy to expand sample diversity and facilitate the training of instruction-following ability. Following this strategy, we can also automatically generate abundant novel instructions to extrapolate to new symbols. To this end, we further evaluate Symbol-LLM by following novel instructions.

\begin{figure}[t]
\large
\centering
\includegraphics[scale=0.52]{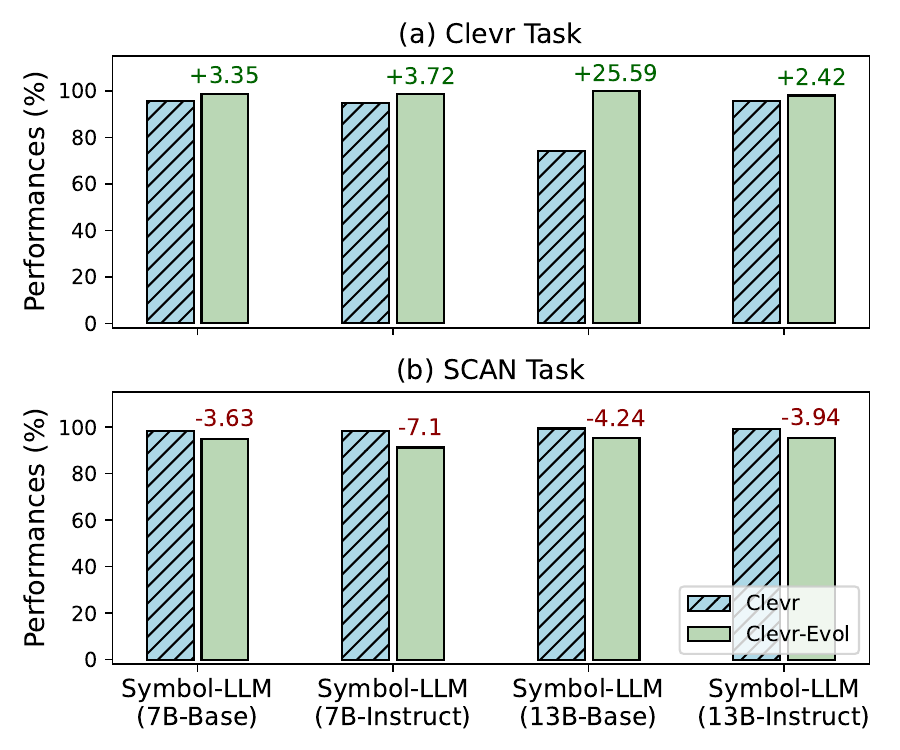}
\caption{Comparisons between original setting and user-defined setting.}
\label{user}
\end{figure}

The experiments are based on \emph{Clevr} and \emph{SCAN} tasks. Applying \emph{symbol-evol} strategy, we obtain \emph{Clevr-evol} and \emph{SCAN-evol} datasets. Evaluation results are presented in Figure~\ref{user}.

From the results, the more complex setting (i.e., green bar) does not induce a significant decrease in model performance. Especially, in the \emph{Clevr} task, Symbol-LLM even does better given the novel instructions. It uncovers that \slm follows the instructions during the reasoning process, instead of merely memorizing the specific symbolic forms.

\section{Analysis: Alignment and Uniformity}
\label{appendix:align_uniform}

Beyond performances on symbolic tasks, it is also required to reveal what leads to superiority. 
Inspired by~\cite{wang2020understanding,yuan-etal-2023-lego}, we extend the ideas of \emph{Alignment} and \emph{Uniformity} to evaluate the model perception of symbolic knowledge. \emph{Alignment}~\footnote{Here, \emph{Alignment} refers to the concept in contrastive learning, but is not related to the alignment technique in LLMs.} is utilized to measure the interrelations between symbolic forms. \emph{Uniformity} quantifies the degree of evenness or uniformity in the distribution of symbolic representations. The concept of a uniform feature distribution is valuable as it encourages a higher information entropy, representing more information retention.

\paragraph{Alignment} Different from the original implementation~\cite{wang2020understanding} which considers positive pairs in the contrastive learning, 
this work takes the symbolic sequences under the same symbolic form as the positive pairs. 
For any symbolic form \emph{X}, their data distributions are referred to as $P_X$. The alignment within \emph{X} can be measured with the following formula:

\begin{equation}
    \label{eq:cal_align}
    \mathcal{L}_{align}(X) = \underset{x_1,x_2\overset{i.i.d}{\sim} P_X}{\mathbb{E}} \left \| f(x_1)-f(x_2) \right \| ^2,
\end{equation}
where $x_1$ and $x_2$ are samples from the specific symbolic form \emph{X}. 
$f(\cdot)$ returns the LLM embeddings of the symbolic sequences. $\left \| \cdot \right \|$ returns the norm of the vector.

In the implementation, we select 16 main symbolic forms and sample 100 symbolic sequences for each form to measure the alignment. $f(\cdot)$ leverages the mean pooling representation of the last hidden states of the LLM. We average all the alignment scores from the 16 symbols to obtain the final one. Notably, we employ logarithmic operations on the \emph{Alignment} loss to reduce scale, without impacting their relative comparison.

\paragraph{Uniformity} Apart from alignment, we also calculate the uniformity of the LLMs on symbolic sequences. The evaluation of the uniformity $\mathcal{L}_{uniform}$ is implemented by the following formula:

\begin{equation}
    \label{eq:cal_uniform}
    \mathcal{L}_{uniform} = \mathrm{log} \underset{x, y\overset{i.i.d}{\sim} P_{data}}{\mathbb{E}} e^{-2\left \| f(x)-f(y) \right \| ^2},
\end{equation}
where the data distribution $P_{data}$ covers all the symbolic sequences. $f(\cdot)$ also utilizes the mean pooling representation of the last hidden states of the LLM.

Leveraging the above definitions, further analysis and comparison on Symbol-LLM are conducted. The item-wise conclusions are listed as follows:

\noindent \textbf{(1) Symbol-LLM optimizes symbol distinctiveness and overall expressiveness in the embedding space (superior \emph{Alignment} and \emph{Uniformity}).}

\begin{figure}[t]
\large
\centering
\includegraphics[scale=0.58]{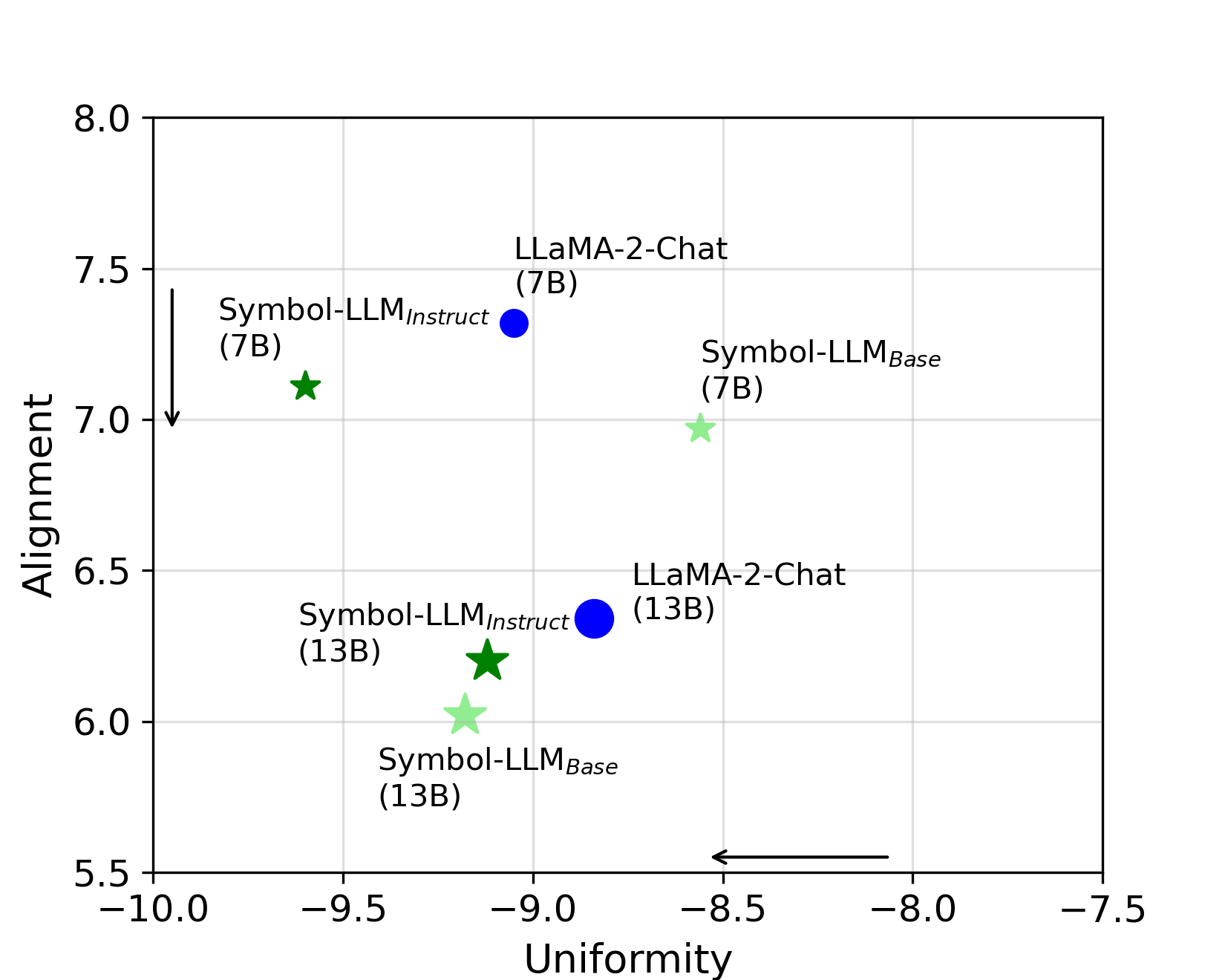}
\caption{Visualization of \emph{Alignment-Uniformity}. Both metrics are inversely related, which means a lower value indicates better performance.}
\label{align_uniform}
\end{figure}

Based on the equation~\ref{eq:cal_align} and~\ref{eq:cal_uniform}, we can assess the proficiency of LLMs in handling symbols. Figure~\ref{align_uniform} presents the visualization of \emph{Alignment-Uniformity}. The x-axis stands for uniformity while the y-axis is the alignment. Both of these metrics are better when kept as small as possible.

From the figure, \slmi models perform consistently better than the original LLaMA-2-Chat models, with obvious merit in \emph{Alignment} and \emph{Uniformity}. It can be regarded as an in-depth explanation for the superior performances on symbolic generation tasks. Further, it witnesses that the two-stage tuning framework actually corrects the weakness of \emph{Uniformity} under 7B settings (\slmi v.s. \slmb).

Both metrics are well optimized with the proposed two-stage tuning framework as well as the symbolic data collection. For \slmb models, though the 7B version witnesses some loss in Uniformity, they consistently achieve superior alignment.

\noindent \textbf{(2) Symbol-LLM excels at capturing symbolic interrelations.}

The above calculation of \emph{Alignment} roughly depicts the similarity among samples under the same symbolic form. To this end, we extend the idea of \emph{Alignment} to measure the interrelation between any two symbolic forms \emph{X} and \emph{Y}. The score $S(X,Y)$ is calculated based on the following formula:

\begin{equation}
    \label{eq:cal_align_2}
    S(X,Y) = \underset{x\sim P_X, y\sim P_Y}{\mathbb{E}} \left \| f(x)-f(y) \right \| ^2,
\end{equation}
where $x$ is one symbolic sample in the form of \emph{X}, while $y$ is one sample in the symbolic form \emph{Y}.
We binarize the scores with the manually defined threshold for a more intuitive illustration. We ensure the same threshold under a fair comparison of the same model size. And the set of thresholds will not affect the overall conclusion.

\begin{figure}[t]
    \centering
    \begin{minipage}[t]{0.49\linewidth}
        \large
        \centering
        \includegraphics[scale=0.27]{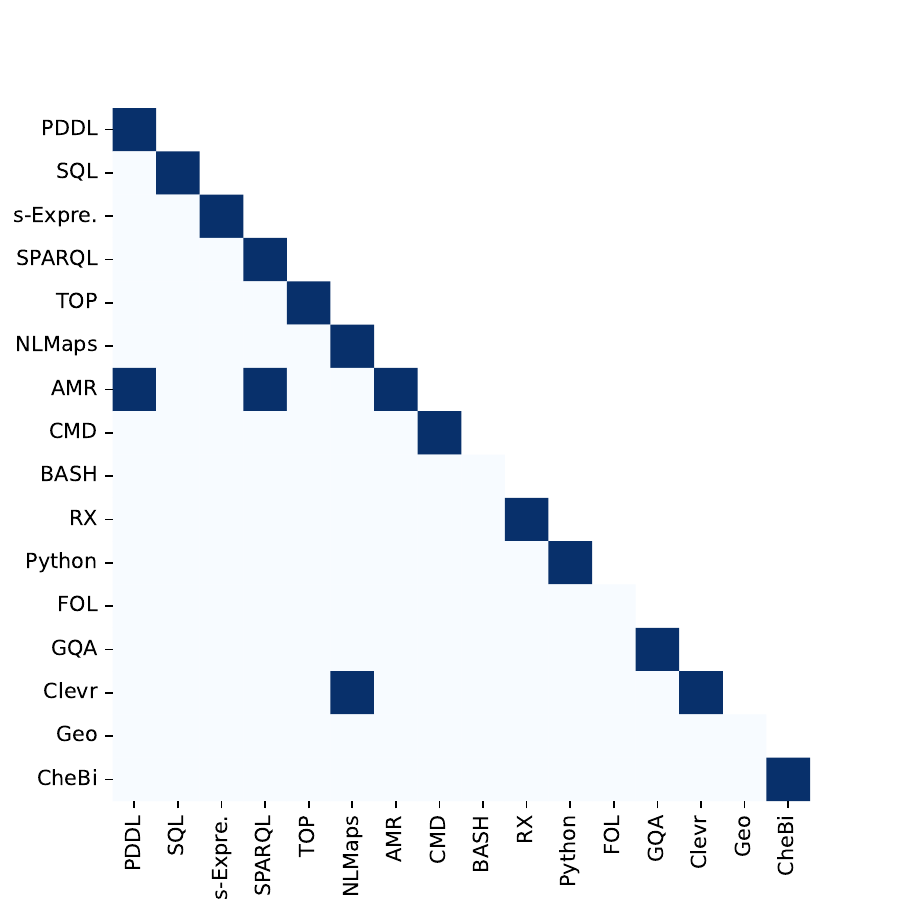}
        \subcaption{LLaMA-2-Chat (7B)}
        \label{align_llama_7b}
    \end{minipage}
    \begin{minipage}[t]{0.49\linewidth}
        \large
        \centering
        \includegraphics[scale=0.27]{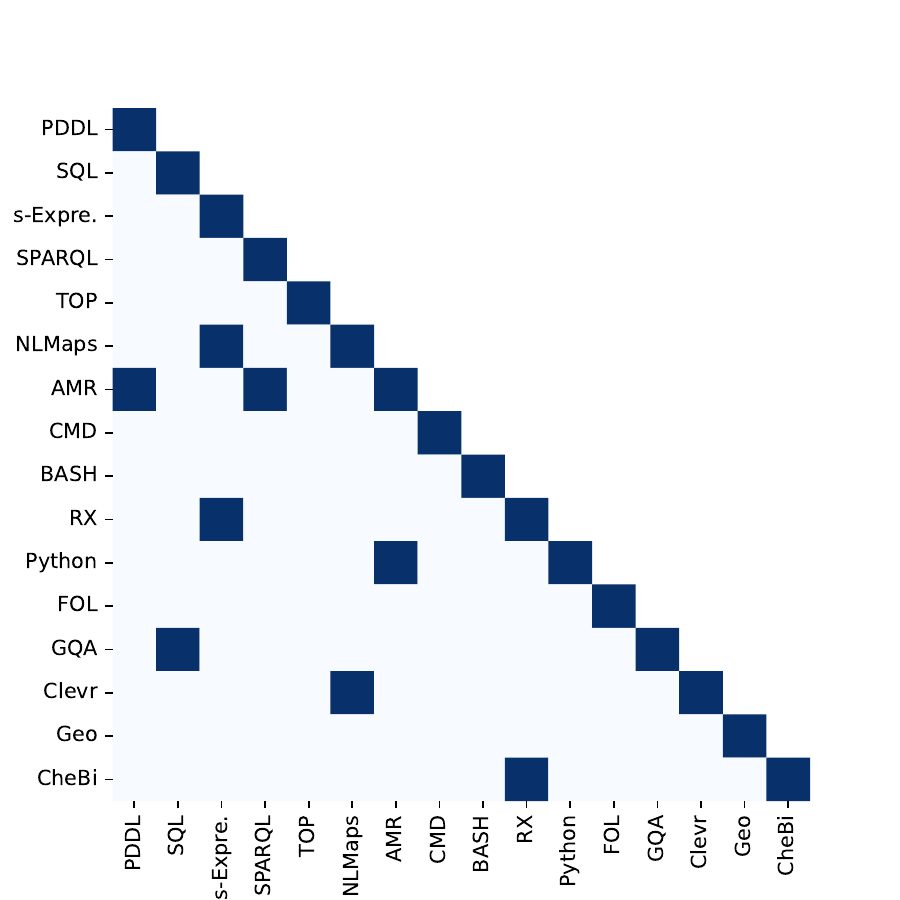}
        \subcaption{\slmi (7B)}
        \label{align_symbolllm_7b}
    \end{minipage}

    \begin{minipage}[t]{0.49\linewidth}
        \large
        \centering
        \includegraphics[scale=0.27]{Figures/align_llama_13b.pdf}
        \subcaption{LLaMA-2-Chat (13B)}
        \label{align_llama_13b}
    \end{minipage}
    \begin{minipage}[t]{0.49\linewidth}
        \large
        \centering
        \includegraphics[scale=0.27]{Figures/align_symbolllm_13b_instruct.pdf}
        \subcaption{\slmi (13B)}
        \label{align_symbolllm_13b}
    \end{minipage}
    \caption{Visualization of the alignment relations between symbols. Dark blue denotes a close relation between two symbols in the representation.}
    \label{alignment}
\end{figure}

The visualization is presented in Figure~\ref{alignment}, where the dark blue denotes the closer relation in the representation while the light one is the opposite. We make comparisons between the original LLaMA-2-Chat models and \slmi models, separately for the size of 7B and 13B.

For the original LLaMA model (Figure~\ref{align_llama_7b} and~\ref{align_llama_13b}), the representations between different symbols exhibit significant sparsity. There are only three pairs of symbolic forms that effectively demonstrate the interrelations in the embedding space, i.e., \emph{AMR-PDDL}, \emph{AMR-SPARQL} and \emph{CLEVR-NLMaps}. Also, under several symbol systems (e.g., \emph{Bash}, \emph{FOL}), the representation space of samples is also very scattered. The above observations demonstrate that previous foundational LLMs (i.e., LLaMA-2-Chat) lack the ability to capture the interrelations among symbolic systems.

In comparison, \slmi series models excel at reflecting the interrelations between symbols. As presented in Figure~\ref{align_symbolllm_7b} and~\ref{align_symbolllm_13b}: 1) Symbols exhibiting potential connections are effectively aligned within the representation space, i.e., \emph{Python-AMR} and \emph{CheBi-RX}. 2) Samples within each symbol are pulled closer together.

Combining the above two observations and analysis, the superior performances of Symbol-LLM on the symbolic generation tasks are sourced from better alignment among symbols in the embedding space as well as the optimized uniformity.

\end{document}